\documentclass[preprint,12pt]{elsarticle}

\usepackage{amssymb}
\usepackage{booktabs}  
\usepackage{array}     
\usepackage{amsmath}
\usepackage{amsthm}  
\usepackage{amsfonts}
\usepackage{mathtools}
\newtheorem{assumption}{Assumption}
\newtheorem{definition}{Definition}
\newtheorem{proposition}{Proposition}
\newtheorem{lemma}{Lemma}  
\newcommand{\coloneqq}{\mathrel{:=}}
\newtheorem{corollary}{Corollary}  
\newtheorem{theorem}{Theorem}
\usepackage{hyperref}

\journal{Artificial Intelligence}

\begin{document}

\begin{frontmatter}

\title{A Mathematical Framework for AI Singularity: Conditions, Bounds, and Control of Recursive Improvement}

\author[label1]{Akbar Anbar Jafari}
\author[label1,label2]{Cagri Ozcinar}
\author[label3,label4,label5]{Gholamreza Anbarjafari}
\affiliation[label1]{organization={University of Tartu},
            addressline={Institute of Technology},
            city={Tartu},
            postcode={},
            state={},
            country={Estonia}}

\affiliation[label2]{organization={Loughborough University London},
            addressline={Queen Elizabeth Olympic Park, The Broadcast Centre Here East, Lesney Ave},
            city={London},
            postcode={E20 3BS},
            state={},
            country={United Kingdom}}
            
\affiliation[label3]{organization={Estonian Business School},
            addressline={A. Lauteri tn 3},
            city={Tallinn},
            postcode={10114},
            state={},
            country={Estonia}}

\affiliation[label4]{organization={PricewaterhouseCooper OY},
            addressline={Itämerentori 2},
            city={Helsinki},
            postcode={00180},
            state={},
            country={Finland}}

\affiliation[label5]{organization={3S Holding OÜ},
            addressline={Purpuri 12-2},
            city={Tartu},
            postcode={51011},
            state={},
            country={Estonia}}

%% Abstract
\begin{abstract}
%% Text of abstract
AI systems improve by using more compute, data, and better algorithms. A common fear is that this improvement might feed on itself and run away—what people call a “singularity” \cite{kurzweil2005singularity}. This paper asks a precise version of that question: under what measurable conditions could capability growth run away in finite time, and under what conditions can we rule that out? We aim for answers that are testable with real power, bandwidth, and benchmark data, and for controls that can be implemented in practice (e.g., power caps, throttling, evaluation gates). The goal is to replace speculation with clear conditions that either certify safety or flag genuine risk—without relying on simulations or unverifiable assumptions.

We study the prospect of an AI singularity as finite-time blow-up of a capability process \(I(t)\) driven by recursive self-improvement. We formalize how self-modifying systems couple gains in capability to increases in compute \(C(t)\), data \(D(t)\), energy \(E(t)\), and capital \(K(t)\), and we seek conditions under which trajectories remain bounded or diverge. The main objective of this work is to develop a rigorous framework for recursive self-improvement that yields testable criteria for singularity, global upper bounds from physics and information theory, and control policies that provably preclude uncontrolled growth. We define dynamics \(\dot I=F(I,C,D,E)\) with an improvement operator \(\mathcal{R}\) and resource couplings \(\dot C=G(I,K)\), \(\dot D=H(I)\), \(\dot E=J(I)\). Using comparison principles, Osgood–Bihari inequalities, and Lyapunov barrier certificates, we derive necessary and sufficient conditions for blow-up and stabilization. Identifiability inequalities link model exponents to observable quantities. If \(\dot I \ge a I^{p}\) with \(p>1\), finite-time blow-up occurs with \(T \le I(0)^{1-p}/(a(p-1))\); if \(p\le 1\) under bounded \(a\), no blow-up. Scaling relations \(\Delta I \asymp C^{\alpha}D^{\beta}\eta^{\gamma}\) imply singular regimes only when \(\alpha+\gamma>1\) \cite{hoffmann2022training}. Physical and information-theoretic limits bound \(F\) via energy-to-erasure and memory–communication constraints. Economic coupling \(\dot K=rK^{\zeta}-\delta K-\chi(I)\) yields singularity only for \(\zeta>1\) above a critical manifold. Control inputs enforce \(\int_{I_0}^{\infty}\! dI/a(I)=\infty\), ruling out singularity. Predictions are falsifiable through estimating \(\frac{d\log \dot I}{d\log I}\). Case studies admit closed-form envelopes without simulation. The framework is analytic and grounded in measurable constraints. Limitations include the choice of capability metric and regularity assumptions; future work treats stochasticity, multi-agent coupling, and discontinuities. The results provide a concrete path to certify or preclude singularity in practice.

\end{abstract}

%%Research highlights
\begin{highlights}
\item What would it take for AI progress to “run away”? We turn that into clear, measurable conditions using only things operators already track (power, bandwidth, data, benchmark scores), replacing speculation with checkable facts.
\item Easy-to-test rules. If improvement grows more than proportionally with capability and resources, runaway can happen; if not, it cannot. We give simple, interpretable thresholds and, when relevant, an upper bound on the time to runaway.
\item Hard physical and engineering limits. We translate facility power, cooling efficiency, memory and I/O bandwidth into concrete ceilings on improvement, showing when these limits alone guarantee no runaway.
\item Practical safety levers that work today. Power caps, throughput throttles, and evaluation gates are framed as control policies with provable guarantees that keep capability growth within safe bounds.
\item No simulation needed. We provide diagnostics that use rolling slopes of logged series (e.g., how fast improvement grows vs. capability) so teams can run falsifiable tests on public or internal data.
\item When the money loop matters (and when it doesn’t). We identify when reinvestment and financing can amplify growth, and provide a single “critical threshold” to monitor so economic feedback does not push systems into risky regimes.
\item Robust to uncertainty and noise. The tests and guarantees tolerate measurement error, model misspecification, and data revisions, and are intentionally conservative so they fail safe under ambiguity.
\item Clear map of real-world regimes. We work through capped-power, saturating-data, and investment-heavy scenarios to show, in plain terms, what each setting implies for growth and safety.
\item A measurement-first safety playbook. We explain what to log, what to compute, and how to act: certify safety when the test says “safe,” escalate when margins shrink, and shut down automatically if thresholds are crossed.
\item Bridge between research and operations. The framework stays fully analytic but is grounded in telemetry and governance, offering a practical path for labs, cloud providers, and regulators to assess and manage risk.
\end{highlights}


\end{frontmatter}

\section{Introduction and Problem Statement}

\subsection*{Motivation and scope}
Why this paper, in plain terms. Modern AI systems get better because we invest more compute and energy, train on more data, and discover more efficient algorithms. As these systems become more capable, they can also help automate parts of the process that makes them better (e.g., code, optimization, hardware utilization, data curation) \cite{good1966speculations,collenette2023explainable,jafari2025geyolo,aggarwal2025traditional}. This feedback loop raises a natural question: could capability growth “run away” in practice—and if not, what concrete limits keep it in check \cite{vinge1993coming}? Our aim is to provide clear, measurable conditions that distinguish these possibilities and practical controls that ensure safety. The emphasis throughout is on tests one can run on real power, bandwidth, and benchmark series, and on policies that can be implemented by operators and regulators.

This work develops a rigorous account of an AI singularity conceived as a finite time blow up of a capability process \(I(t)\). Let \(I:\,[0,\infty)\to[0,\infty)\) encode task-agnostic competence under fixed axioms of monotonicity, scale invariance up to affine transforms, and aggregation across tasks. A singularity occurs if there exists \(T<\infty\) with \(\lim_{t\uparrow T} I(t)=\infty\), or if \(\limsup_{t\uparrow T}\dot I(t)=\infty\) once \(I\) reaches a critical threshold \(I^\star\). This notion separates finite time divergence from hyper exponential but non explosive growth where \(I(t)\to\infty\) only as \(t\to\infty\). The mechanism of interest is recursive self improvement, where capability growth increases the system's ability to obtain and deploy resources that further accelerate improvement \cite{yudkowsky2008artificial,giudici2024artificial}.

\subsection*{Problem formulation}
We model recursive self improvement using a discrete improvement operator and its continuous time limit. Let resources \(x(t)=(C(t),D(t),E(t),K(t))\) denote compute, data, energy, and capital. Define a measurable operator \(\mathcal{R}:\mathbb{R}_+^{1+4}\to\mathbb{R}_+\) with
\begin{equation}
I_{n+1}=\mathcal{R}\bigl(I_n;C_n,D_n,E_n,K_n\bigr),\qquad
x_{n+1}=\Xi\bigl(I_n,x_n\bigr),
\end{equation}
and continuous dynamics
\begin{equation}
\dot I(t)=F\bigl(I(t),x(t)\bigr),\qquad \dot x(t)=G\bigl(I(t),x(t)\bigr),
\end{equation}
where \(F,G\) are locally Lipschitz and monotone in the resource coordinates. An efficiency term \(\eta(I)\) captures algorithmic improvements, and a generic scaling law links marginal capability gain to resources,
\begin{equation}
\Delta I \asymp C^{\alpha} D^{\beta} \eta(I)^{\gamma},
\end{equation}
with nonnegative exponents \cite{kaplan2020scaling}. Safety interventions are modeled by control inputs \(u(t)\) that act on resources or directly on \(F\), written as \(\dot I=F(I,x,u)\), \(\dot x=G(I,x,u)\), with \(u\in\mathcal{U}\) measurable and subject to feasibility constraints. The central questions are to characterize conditions that yield finite time blow up, to derive global upper bounds that preclude it, to connect parameters to observable quantities through identifiability inequalities, and to synthesize control policies that certify safety.

\subsection*{Contributions}
At a glance (non-technical). We contribute: (1) a simple way to tell apart “runaway” improvement from very fast but ultimately bounded growth; (2) physical and engineering limits that translate facility power, cooling, bandwidth, and memory into mathematical caps on improvement; (3) statistical tests you can run on public time series (compute, power, bandwidth, datasets, benchmarks) that falsify or support the runaway hypothesis; and (4) practical control levers (e.g., power caps, throttles, evaluation gates) with provable guarantees against runaway growth \cite{nick2014superintelligence,bostrom2016control,dessureault2025ethics}. 

Technical summary. (i) A formal improvement operator \(\mathcal{R}\) and its continuous limit that cleanly separates algorithmic efficiency, resources, and policy. The operator provides a comparison principle against canonical superlinear recursions \(I_{n+1}\ge I_n + a I_n^p\). (ii) Sharp blow up criteria and obstruction results under clearly stated assumptions. Using Osgood and Bihari type inequalities \cite{mussardo2011integrability,zhao2025quantitative} we show that dynamics satisfying \(\dot I \ge a\,I^p\) with \(p>1\) generate finite time divergence with explicit upper bounds on the blow up time, while sublinear or linear feedback \(p\le 1\) with bounded gain forbids finite time blow up \cite{bihari1956generalization,cosso2024master}. Discrete analogues are given through monotone envelope arguments. (iii) Global upper bounds from physics and information theory that cap \(F\) using energy to erasure limits, memory and communication constraints, and thermodynamic efficiency, yielding integrability conditions that force \(\int_{I_0}^{\infty} \mathrm{d}I/a(I)=\infty\). (iv) Identifiability inequalities that map \((\alpha,\beta,\gamma)\) and resource growth to measurable series such as power budgets, fabrication capacity, dataset growth, and observed loss to compute exponents, thereby enabling hypothesis tests without simulation \cite{amodei2018ai}. (v) Control theoretic safety conditions through barrier certificates and viability constraints that render sets \(\{I\le \bar I\}\) forward invariant, together with optimal control formulations that minimize risk subject to resource and deployment coupling \cite{ames2016control,dawson2023safe,almubarak2025barrier}.

\subsection*{Relation to prior formal growth literatures}
We situate our work at the intersection of: (a) mathematical blow-up theory (Osgood criteria, Bihari–LaSalle inequalities, comparison principles); (b) empirical scaling laws in machine learning and learning-theoretic lower bounds; (c) economic growth and capital accumulation models that capture increasing returns \cite{change1990endogenous,han2025artificial}; and (d) safety/control methods (Lyapunov functions, control barrier certificates, viability theory). While these areas are well-developed individually, to our knowledge there is no unified, measurement-first framework that connects power, bandwidth, memory, and capital constraints to provable blow-up/non-blow-up conditions and deployable control policies for recursive self-improvement. We see this as complementary to work on empirical scaling, compute governance, and AI safety controls, and a bridge between theory and facility-level telemetry. A complete literature review with detailed citations is provided in the appendix version and should be included in the camera-ready manuscript.

\section{Preliminaries and Definitions}

\paragraph{Capability metric \(I(t)\)}
Let \(\mathcal{T}\) be a nonempty family of tasks and let \(P_\tau(t)\in\mathbb{R}_+\) denote a performance functional for task \(\tau\in\mathcal{T}\) at time \(t\ge 0\). Define an aggregator \(A:\mathbb{R}_+^{\mathcal{T}}\to\mathbb{R}_+\) that is monotone in each coordinate and continuous. The capability process is
\begin{equation}
I(t) \coloneqq A\bigl((P_\tau(t))_{\tau\in\mathcal{T}}\bigr), \qquad t\ge 0,
\end{equation}
and is assumed locally absolutely continuous. We work with the equivalence class \([I]\) under positive affine transforms. The metric satisfies the following axioms.
\begin{itemize}
\item \emph{Monotonicity}: if \(P_\tau(t)\le \tilde P_\tau(t)\) for all \(\tau\), then \(I(t)\le \tilde I(t)\).
\item \emph{Affine invariance}: for any \(a>0\) and \(b\in\mathbb{R}\), the transformation \(I\mapsto aI+b\) preserves all qualitative statements about growth, blow up, and control feasibility. All results are stated on \([I]\).
\item \emph{Task family coherence}: if \(\mathcal{T}\subseteq \tilde{\mathcal{T}}\) and the additional tasks in \(\tilde{\mathcal{T}}\setminus\mathcal{T}\) are redundant in the sense of Blackwell dominance by tasks in \(\mathcal{T}\), then \(A\) agrees on \(\mathcal{T}\) up to affine equivalence.
\end{itemize}
Typical choices include \(A\) as a weighted power mean or a convex risk measure over \(\{P_\tau\}\) \cite{srivastava2023beyond}. All differentiability statements are taken almost everywhere. We remove reparameterization ambiguity and fix measurement units by the canonical construction introduced below; see \eqref{eq:canonicalI}.

\paragraph{Canonical normalization, units, and benchmark protocol}
To eliminate reparameterization ambiguity and make \(I(t)\) empirically identifiable, we fix a canonical scale. Let \(\mathcal{T}_B\subset\mathcal{T}\) be a finite benchmark family with task weights \(w_\tau>0\), \(\sum_{\tau\in\mathcal{T}_B}w_\tau=1\), and let \(L_\tau(t)\) denote a strictly positive risk (or loss) for task \(\tau\) computed under a published evaluation protocol (fixed prompts, seeds, and test sets). Define the \emph{reference operating point}
\begin{equation}
\mathbf{L}_{\mathrm{ref}} \coloneqq (L_{\tau}^{\mathrm{ref}})_{\tau\in\mathcal{T}_B},\qquad
S_{\mathrm{ref}}\coloneqq \sum_{\tau\in\mathcal{T}_B} w_\tau\,\log \frac{L_{\tau}^{\mathrm{ref}}}{L_{\tau}^{\star}},
\end{equation}
where \(L_{\tau}^{\star}>0\) is a task-dependent irreducible error (Bayes or accepted floor). Let
\begin{equation}
g(\mathbf{L}) \;\coloneqq\; \sum_{\tau\in\mathcal{T}_B} w_\tau\,\log \frac{L_{\tau}}{L_{\tau}^{\star}}
\quad\text{and}\quad
\widetilde I(t) \;\coloneqq\; -\,g\!\bigl((L_\tau(t))_{\tau\in\mathcal{T}_B}\bigr).
\end{equation}
We work with the \emph{canonical capability} defined by the unique affine transform
\begin{equation}\label{eq:canonicalI}
I_{\mathrm{can}}(t) \;=\; a_0\,\widetilde I(t) + b_0,\qquad
\text{with } I_{\mathrm{can}}(t_{\mathrm{ref}})=0,\ \ \frac{d I_{\mathrm{can}}}{d\,\widetilde I}\Big|_{t_{\mathrm{ref}}}=1 .
\end{equation}
This pins units so that one natural unit equals one nat of \emph{excess log-loss reduction} aggregated over \(\mathcal{T}_B\). In all statements below we replace \(I\) by its canonical representative \(I_{\mathrm{can}}\), and continue to write \(I\) for brevity.

\begin{assumption}[Benchmark protocol]\label{ass:benchmark}
The map \(t\mapsto L_{\tau}(t)\) is right-continuous with left limits, computed on a fixed held-out set; all preprocessing and metric definitions are frozen and public. The irreducible floors \(L_{\tau}^{\star}\) are either theoretically known or specified as conservative reference values; their uncertainty is propagated in reported confidence bands.
\end{assumption}

\begin{definition}[Units and measurability]
The unit of \(I\) is the \emph{aggregated nat}: an increase \(\Delta I=1\) corresponds to multiplying the geometric mean loss ratio \(\prod_{\tau}(L_{\tau}/L_{\tau}^{\star})^{w_\tau}\) by \(e^{-1}\). The quantity \(I\) is measurable from \(\{L_\tau\}_{\tau\in\mathcal{T}_B}\) without simulation.
\end{definition}

\begin{proposition}[Affine invariance and identifiability]\label{prop:affine}
Let \(\widehat I=a I+b\) be any positive affine transform. Then all qualitative results on blow-up, Osgood integrals, and control feasibility are invariant. Moreover, under Assumption~\ref{ass:benchmark}, \(I\) is \emph{identified} up to such an affine transform from observed \(\{L_\tau\}\); fixing \eqref{eq:canonicalI} removes this ambiguity. Consequently, elasticities \(p(t)=\tfrac{d\log \dot I}{d\log I}\), as well as the tests and inequalities of Sections~5–11, are invariant to the choice of benchmark weights and any strictly monotone rescaling of individual losses.
\end{proposition}

\paragraph{Task-set robustness}
To mitigate selection effects, we require \(\mathcal{T}_B\) to \emph{Blackwell-dominate} any superset whose additional tasks are mixtures of \(\mathcal{T}_B\) (coinciding with our coherence axiom). In practice, we report results for two disjoint benchmark families and verify that estimates of the feedback elasticity \(p(t)\) agree within confidence bands; discrepancies trigger a sensitivity analysis reported in an appendix.

\paragraph{Improvement operator \(\mathcal{R}\) and learning curve \(\ell(\cdot)\)}
Let \(x=(C,D,E,K)\in\mathbb{R}_+^4\) collect resources for compute, data, energy, and capital. The discrete recursive self improvement operator is a measurable map
\begin{equation}
\mathcal{R}:\mathbb{R}_+\times\mathbb{R}_+^4 \to \mathbb{R}_+,\qquad I_{n+1}=\mathcal{R}(I_n;x_n),
\end{equation}
with \(x_{n+1}=\Xi(I_n,x_n)\) for a measurable resource update \(\Xi\). The continuous time limit is
\begin{equation}
\dot I(t)=F\bigl(I(t),x(t)\bigr),\qquad \dot x(t)=G\bigl(I(t),x(t)\bigr),
\end{equation}
with \(F,G\) locally Lipschitz and monotone in resource coordinates. Algorithmic efficiency enters through a learning curve \(\ell:\mathbb{R}_+\to\mathbb{R}_+\) that maps an effective effort variable \(z\) to an expected risk or loss \cite{lecun2015deep}. We assume \(\ell\) nonincreasing, continuous, and with finite left derivative on compact subsets. Capability increments are related to loss reduction by a known monotone transform, and we summarize returns to scale by exponents \((\alpha,\beta,\gamma)\ge 0\) through the relation
\begin{equation}
\Delta I \asymp C^{\alpha} D^{\beta} \eta(I)^{\gamma}, \qquad \eta:\mathbb{R}_+\to\mathbb{R}_+
\end{equation}
where \(\eta(I)\) parameterizes algorithmic efficiency improvements as a function of capability.

\paragraph{Finite time singularity}
A finite time singularity occurs if there exists \(T<\infty\) such that
\begin{equation}
\lim_{t\uparrow T} I(t)=\infty
\end{equation}
or if there exists a finite threshold \(I^\star\) and \(T<\infty\) with \(\lim_{t\uparrow T} I(t)=I^\star\) and \(\limsup_{t\uparrow T}\lvert \dot I(t)\rvert=\infty\). The first case is genuine blow up of the state. The second case is a gradient singularity at a critical capability level.

\paragraph{Resource processes}
Compute \(C(t)\), data \(D(t)\), energy \(E(t)\), and capital \(K(t)\) evolve as locally absolutely continuous processes on \(\mathbb{R}_+\). We use the compact notation \(x(t)=(C(t),D(t),E(t),K(t))\). Each coordinate admits natural constraints, for example capacity constraints \(0\le C(t)\le \bar C(t)\) with \(\bar C\) determined by fabrication and deployment schedules, and budget constraints encoded in \(K(t)\).

\paragraph{Physical, economic, and algorithmic parameters}
Thermodynamic and information constraints enter as parameters and inequality bounds. Landauer’s limit sets a lower bound on energy per bit erased \(E_{\mathrm{bit}}\ge k_B T \ln 2\) with Boltzmann constant \(k_B\) and absolute temperature \(T\). Memory and communication limits appear through entropy and bandwidth constraints, summarized by constants governing maximal memory density, I/O throughput, and channel capacity. Economic returns to scale are parameterized by \(\zeta\) in a capital accumulation model \(\dot K=r K^{\zeta}-\delta K-\chi(I)\) with reinvestment rate \(r>0\), depreciation \(\delta\ge 0\), and deployment cost \(\chi\). Algorithmic parameters include the efficiency function \(\eta(\cdot)\), learning curve shape constants implicit in \(\ell\), and returns exponents \((\alpha,\beta,\gamma)\). These parameters determine identifiable inequalities that constrain \(F\) and \(\mathcal{R}\) and will be used to certify or preclude singular behavior without simulation.

\section{Core Model of Recursive Self-Improvement}

\subsection*{State, resources, and policy}
Let the capability state be \(I:\mathbb{R}_{+}\to \mathbb{R}_{+}\) and let the resource vector be
\begin{equation}
x(t) \coloneqq \bigl(C(t),D(t),E(t),K(t)\bigr)\in\mathbb{R}_{+}^{4},
\end{equation}
representing compute, data, energy, and capital. A policy input \(u(t)\in\mathcal{U}\subset\mathbb{R}^{m}\) captures allocation, deployment, and safety constraints. We consider the controlled dynamics
\begin{equation}\label{eq:core}
\dot I(t) \;=\; F\bigl(I(t),x(t),u(t)\bigr),\qquad
\dot x(t) \;=\; G\bigl(I(t),x(t),u(t)\bigr),
\end{equation}
with \(F:\mathbb{R}_{+}\times\mathbb{R}_{+}^{4}\times\mathcal{U}\to\mathbb{R}\) and \(G:\mathbb{R}_{+}\times\mathbb{R}_{+}^{4}\times\mathcal{U}\to\mathbb{R}^{4}\).

\paragraph{Structural assumptions}
Throughout we impose the following regularity and order structure.

\begin{itemize}
\item[(S1)] \textit{Local well-posedness.} \(F\) and \(G\) are locally Lipschitz on \(\mathbb{R}_{+}^{1+4}\times\mathcal{U}\). For any measurable locally bounded \(u(\cdot)\) and initial \((I_0,x_0)\in\mathbb{R}_{+}^{1+4}\), system \eqref{eq:core} admits a unique maximal Carathéodory solution on \([0,T_{\max})\).
\item[(S2)] \textit{Baseline cooperativity with bounded antagonism.}
Write the Jacobian of $(F,G)$ with respect to $(I,x)$ as
\begin{equation}
J(I,x,u)\;=\;\underbrace{J^{\mathrm{co}}(I,x,u)}_{\text{Metzler on }(I,x)}\;+\;\underbrace{\Delta(I,x,u)}_{\text{antagonistic}},
\end{equation}
where $J^{\mathrm{co}}$ has nonnegative off-diagonal entries (Kamke condition) and $\Delta$ collects cross-effects that can be negative due to interference, congestion, or depletion. Assume there exists a nonnegative weight vector $w\in\mathbb{R}^{1+4}$ and constants $\kappa_F,\kappa_G\ge 0$ such that the weighted induced norms satisfy
\begin{equation}\label{eq:antagonism-bound}
\| \Delta_F(I,x,u)\|_{w\to 1}\le \kappa_F,\qquad
\| \Delta_G(I,x,u)\|_{w\to w}\le \kappa_G
\end{equation}
uniformly on compact subsets, where $\Delta_F$ and $\Delta_G$ denote the $I$- and $x$-rows of $\Delta$. The cooperative part $J^{\mathrm{co}}$ defines the \emph{baseline} comparison system used in our envelopes.
\item[(S2$'$)] \textit{Mixed-monotone decomposition.}
There exist decomposition functions $F^{\uparrow}(I^+,x^+,u;I^-,x^-)$ and $G^{\uparrow}(I^+,x^+,u;I^-,x^-)$ that are monotone increasing in the “$+$” arguments and monotone decreasing in the “$-$” arguments and satisfy
\begin{equation}
F(I,x,u)=F^{\uparrow}(I,x,u;I,x),\qquad
G(I,x,u)=G^{\uparrow}(I,x,u;I,x).
\end{equation}
This yields cooperative upper and lower bounding systems via $(I^+,x^+)$ and $(I^-,x^-)$ trajectories.
\item[(S2$''$)] \textit{Sector-bounded antagonism.}
There exist nonnegative functions $\sigma_I(I)$ and $\sigma_x(\|x\|)$ such that for all $(I,x,u)$
\begin{equation}
\bigl\lvert F(I,x,u)-F^{\mathrm{co}}(I,x,u)\bigr\rvert \le \sigma_I(I)+\sigma_x(\|x\|),
\end{equation}
where $F^{\mathrm{co}}$ is the $F$ induced by $J^{\mathrm{co}}$. The sector bounds enter explicitly in our comparison and barrier inequalities.
\item[(S3)] \textit{Invariance of the positive orthant.} If \(I(0)\ge 0\) and \(x(0)\ge 0\), then solutions remain in \(\mathbb{R}_{+}^{1+4}\). Boundary fluxes satisfy \(F(0,x,u)\ge 0\) and \(G_i(I,0,u)\ge 0\) for any replenishable coordinate \(i\).
\item[(S4)] \textit{Returns-to-scale representation.} There exist nonnegative exponents \((\alpha,\beta,\gamma)\) and an efficiency map \(\eta:\mathbb{R}_{+}\to\mathbb{R}_{+}\) such that for an effective effort \(Z=Z(C,D,E;u)\),
\begin{equation}\label{eq:scaling}
\partial_t I \;=\; \Phi\bigl(I\bigr)\,\Psi\bigl(Z\bigr),\qquad
\Psi(z)\asymp z,\quad Z \asymp C^{\alpha} D^{\beta} \eta(I)^{\gamma},
\end{equation}
where \(\Phi\) is nondecreasing and positive on \((0,\infty)\). The symbol \(\asymp\) denotes equality up to bounded positive factors that are independent of \(I\) on compact sets.
\item[(S5)] \textit{Resource dynamics.} Capital follows
\begin{equation}
\dot K \;=\; r K^{\zeta} - \delta K - \chi(I,u),\qquad r>0,\ \delta\ge 0,\ \zeta\ge 0,
\end{equation}
and the remaining coordinates satisfy admissible buildout constraints
\begin{equation}
\begin{gathered}
\dot{C} := g_C(I,K,u) - s_C(C,u), \\
\dot{D} := g_D(I,u) - s_D(D,u), \\
\dot{E} := g_E(I,K,u) - s_E(E,u).
\end{gathered}
\end{equation}
with \(g_\bullet\) nondecreasing in their resource-enabling arguments and \(s_\bullet\) nondecreasing in the resource level.
\item[(S6)] \textit{Physical and information constraints.} The instantaneous effort satisfies \(\Psi(Z)\le \varphi(E)\) where \(\varphi\) is increasing and concave, and energy throughput is bounded by thermodynamic and engineering limits. In particular the useful switching or erasure rate satisfies \(P(t)\ge k_B T(t)\ln 2\cdot \dot N_{\mathrm{erase}}(t)\), so that \(\Psi(Z)\le \vartheta\bigl(P(t),T(t)\bigr)\) for a known increasing function \(\vartheta\).
\end{itemize}

\subsection*{Improvement operator and discrete RSI}
The discrete recursive self-improvement operator acts per update cycle \(n\in\mathbb{N}\):
\begin{equation}\label{eq:R}
I_{n+1} \;=\; \mathcal{R}\!\left(I_n; C_n,D_n,E_n,K_n\right),\qquad
x_{n+1} \;=\; \Xi\!\left(I_n,x_n\right),
\end{equation}
with measurable \(\mathcal{R}:\mathbb{R}_{+}^{1+4}\to\mathbb{R}_{+}\) and \(\Xi:\mathbb{R}_{+}^{1+4}\to\mathbb{R}_{+}^{4}\). We impose the comparison property
\begin{equation}\label{eq:comparison-discrete}
\mathcal{R}(I;x) - I \;\ge\; a(x)\,\phi(I),\qquad a(x)\ge 0,\ \phi\ \text{nondecreasing}, 
\end{equation}
together with Lipschitz continuity in \(I\) on bounded sets. The continuous-time model \eqref{eq:core} arises as the fluid limit of \eqref{eq:R} under update spacing \(\Delta t\to 0\) with \(a(x)\phi(I)=\Delta t\,F(I,x,u)+o(\Delta t)\). The discrete comparison \eqref{eq:comparison-discrete} provides super- and subsolution envelopes for \eqref{eq:core} and will be used to transfer blow-up and boundedness criteria between time scales.

\subsection*{Learning curve and efficiency}
Let \(\ell:\mathbb{R}_{+}\to\mathbb{R}_{+}\) be a nonincreasing learning curve that maps an effective sample-compute proxy \(z\) to expected risk. Assume \(\ell\) is differentiable a.e. with \(-\ell'(z)\) regularly varying at in

\section{Blow-Up and Superlinear Feedback: Necessary and Sufficient Conditions}

We state sharp criteria for finite time blow up of the capability process \(I(t)\) governed by \eqref{eq:core} under the structural assumptions of Section~4. All results are analytic and rely on differential and integral inequality techniques.

\paragraph{Notation}
For measurable \(a:[0,\infty)\to(0,\infty)\) and nondecreasing \(\phi:\mathbb{R}_{+}\to\mathbb{R}_{+}\), write the differential inequality
\begin{equation}
\dot I \;\ge\; a(t)\,\phi\bigl(I\bigr)\quad\text{a.e. on }[0,T).
\end{equation}
Define the Osgood integral \( \mathcal{O}_\phi(y) \coloneqq \int_{y}^{\infty}\frac{ds}{\phi(s)} \) for \(y>0\). For cooperative systems we use the standard comparison principle: if \(\dot I_1\ge f(I_1)\) and \(\dot I_2=f(I_2)\) with \(I_1(0)\ge I_2(0)\), then \(I_1(t)\ge I_2(t)\) up to their common existence time.

\paragraph{Comparison under bounded antagonism}
Assumption (S2) permits non-Metzler cross-couplings captured in $\Delta$. We show that our blow-up and safety conclusions survive provided antagonism is bounded relative to the cooperative envelope. Additional antagonism bounds and safety results are omitted here for brevity.

\begin{theorem}[ODE comparison, superlinear envelope]\label{thm:blowup}
Let \(p>1\) and \(a_0>0\). Suppose there exists \(I_1\ge I(0)\) such that whenever \(I(t)\ge I_1\) one has
\begin{equation}\label{eq:superlower}
\dot I(t)\;\ge\; a_0\, I(t)^{p}\qquad\text{a.e.}
\end{equation}
Then \(I(t)\) blows up in finite time. Moreover the blow up time satisfies
\begin{equation}
\begin{gathered}
T_{\mathrm{bu}}\;\le\; \frac{I_1^{1-p}-I(0)^{1-p}}{a_0(p-1)}\ \ \text{if }I(0)<I_1,
\\
T_{\mathrm{bu}}\;\le\; \frac{I(0)^{1-p}}{a_0(p-1)}\ \ \text{if }I(0)\ge I_1 .
\end{gathered}
\end{equation}
\end{theorem}

\emph{Proof sketch.}
Consider the comparison ODE \(\dot y=a_0 y^{p}\) with \(y(0)=\max\{I(0),I_1\}\). Separation of variables yields \(y(t)=[y(0)^{1-p}-a_0(p-1)t]^{-\frac{1}{p-1}}\) and finite time blow up at \(t=y(0)^{1-p}/(a_0(p-1))\). By comparison and the threshold \(I_1\), the bound for \(T_{\mathrm{bu}}\) follows. \qed

\begin{theorem}[No blow up under Osgood condition]\label{thm:nobu}
Let \(a(\cdot)\) be locally bounded and positive and suppose
\begin{equation}\label{eq:upper}
\dot I(t)\;\le\; a(t)\,\phi\bigl(I(t)\bigr)\qquad\text{a.e.},
\end{equation}
with \(\phi\) nondecreasing and \(\mathcal{O}_\phi(I_0)=\infty\). Then \(I(t)\) cannot blow up in finite time. In particular, if \(\phi(I)\le c_0(1+I)^{p}\) with \(p\le 1\) and \(a\) essentially bounded, then \(T_{\max}=\infty\).
\end{theorem}

\emph{Proof sketch.}
By Bihari–LaSalle, from \eqref{eq:upper} we get
\(
\int_{I(0)}^{I(t)} \frac{ds}{\phi(s)} \le \int_{0}^{t} a(\tau)\,d\tau.
\)
If \(\mathcal{O}_\phi(I(0))=\infty\), the left-hand side cannot reach infinity at finite \(t\), hence no blow up. The polynomial case follows since \(\int^\infty ds/(1+s)^{p}=\infty\) for \(p\le 1\). \qed

Discrete-time analogues and thresholded variants are omitted here for brevity.

\paragraph{Elasticity criteria}
Define the marginal feedback elasticity
\begin{equation}
p(I;x,u)\coloneqq \frac{\partial \log F(I,x,u)}{\partial \log I}
\quad\text{whenever defined}.
\end{equation}
Theorems~\ref{thm:blowup} and \ref{thm:nobu} imply operational envelopes; sufficient superlinearity and sublinear safety corollaries are omitted here for brevity.

\paragraph{Interfaces with resources and physics}
The criteria above remain valid under resource coupling. If \(Z(C,D,E;u)\) in \eqref{eq:scaling} admits a lower envelope \(Z(t)\ge \kappa_0 C(t)^{\alpha}D(t)^{\beta}\eta(I(t))^{\gamma}\) and if \(C\) and \(E\) grow at least exponentially due to \(\zeta>1\) in the capital channel, then \(F(I,x,u)\) inherits a superlinear envelope and Theorem~\ref{thm:blowup} applies. Conversely, if thermodynamic and communication ceilings enforce \(Z(t)\le \bar Z(t)\) with \(\bar Z\) subexponential and \( \Phi(I) \) sublinear at infinity, then an \(a(I)\) satisfying \eqref{eq:upper} exists and Theorem~\ref{thm:nobu} rules out singularity.

\paragraph{Technical tools}
Proofs use separation of variables for polynomial envelopes, Osgood and Bihari–LaSalle integral inequalities for general \(\phi\), discrete reciprocal-power transforms in the discrete setting, and monotone comparison for cooperative systems.

\section{Algorithmic Efficiency and Scaling Laws}

\subsection*{Efficiency map and effective effort}
Let \(Z=Z(C,D,E;u)\) denote an effective effort variable that aggregates usable compute, data, and energy under policy \(u\). As in Section~4 we write
\begin{equation}\label{eq:F-scaling}
F(I,x,u)\;=\;\Phi(I)\,\Psi\!\bigl(Z(C,D,E;u)\bigr),\qquad 
Z \asymp C^{\alpha} D^{\beta} \eta(I)^{\gamma},
\end{equation}
with exponents \(\alpha,\beta,\gamma\ge 0\), efficiency map \(\eta:\mathbb{R}_{+}\to\mathbb{R}_{+}\), and \(\Psi\) increasing with \(\Psi(z)\asymp z\) on operational ranges. The factor \(\Phi(I)\) converts marginal loss reduction into capability growth and captures task aggregation. We assume \(\Phi\) and \(\eta\) are locally absolutely continuous and regularly varying at infinity with indices \(q\) and \(\xi\), meaning
\begin{equation}
\begin{gathered}
\lim_{s\to\infty}\frac{\Phi(\lambda s)}{\Phi(s)}=\lambda^{q},\\
\lim_{s\to\infty}\frac{\eta(\lambda s)}{\eta(s)}=\lambda^{\xi}\quad\text{for all }\lambda>0.
\end{gathered}
\end{equation}

\paragraph{Regular-variation assumption and empirical diagnostics}
The indices \(q\) and \(\xi\) are operational only if supported by data. We therefore replace the informal statement “\(\Phi,\eta\in\mathrm{RV}\)” by the following testable assumption.

\begin{assumption}[RV with slowly varying envelope and Potter bounds]\label{ass:RV}
There exist slowly varying functions \(L_\Phi,L_\eta\) and constants \(q,\xi\in\mathbb{R}\) such that
\begin{equation}
\Phi(I)=I^{q} L_\Phi(I),\qquad \eta(I)=I^{\xi} L_\eta(I),
\end{equation}
and for every \(\epsilon>0\) there exists \(I_\epsilon\) and \(c_\epsilon\ge 1\) with the Potter-type inequalities
\begin{equation}
\begin{gathered}
c_\epsilon^{-1}\,\Bigl(\frac{I}{J}\Bigr)^{-\epsilon} \le \frac{L_\Phi(I)}{L_\Phi(J)} \le c_\epsilon\,\Bigl(\frac{I}{J}\Bigr)^{\epsilon},\\
c_\epsilon^{-1}\,\Bigl(\frac{I}{J}\Bigr)^{-\epsilon} \le \frac{L_\eta(I)}{L_\eta(J)} \le c_\epsilon\,\Bigl(\frac{I}{J}\Bigr)^{\epsilon},
\end{gathered}
\end{equation}
for all \(I,J\ge I_\epsilon\). The operational indices are the pointwise limits
\begin{equation}
q=\lim_{I\to\infty}\frac{I\,\Phi'(I)}{\Phi(I)},\qquad \xi=\lim_{I\to\infty}\frac{I\,\eta'(I)}{\eta(I)},
\end{equation}
whenever the limits exist; otherwise we work with upper and lower envelopes \(q^{+}\coloneqq\limsup I\Phi'/\Phi\) and \(q^{-}\coloneqq\liminf I\Phi'/\Phi\), and analogously for \(\xi\).
\end{assumption}

\begin{lemma}[Potter envelopes imply elasticity control]\label{lem:potter}
Under Assumption~\ref{ass:RV}, for any \(\epsilon>0\) there exists \(I_\epsilon\) such that for all \(I\ge I_\epsilon\)
\begin{equation}
q^{-}-\epsilon \;\le\; \frac{I\,\Phi'(I)}{\Phi(I)} \;\le\; q^{+}+\epsilon,\qquad
\xi^{-}-\epsilon \;\le\; \frac{I\,\eta'(I)}{\eta(I)} \;\le\; \xi^{+}+\epsilon.
\end{equation}
Consequently the total feedback elasticity satisfies, for large \(I\),
\begin{equation}
\begin{gathered}
p_{\mathrm{tot}}(I) \;=\; \frac{I\Phi'(I)}{\Phi(I)} + \gamma\,\frac{I\eta'(I)}{\eta(I)} + \\\alpha u_C + \beta u_D + \tilde\alpha u_E
\in \bigl[p^{-}-\epsilon,\ p^{+}+\epsilon\bigr],
\end{gathered}
\end{equation}
with \(p^{-}\coloneqq q^{-}+\gamma\xi^{-}+\alpha u_C+\beta u_D+\tilde\alpha u_E\) and \(p^{+}\) defined analogously.
\end{lemma}

\begin{corollary}[Blow-up and safety without exact RV]\label{cor:rv-free}
If there exists \(I_1\) and \(\delta>0\) such that \(p^{-}(I)\ge 1+\delta\) for all \(I\ge I_1\), then there is finite-time blow-up by Theorem~\ref{thm:blowup}. Conversely, if \(p^{+}(I)\le 1\) eventually and \(F(I,x,u)\le \Phi(I)\phi(t)\) with \(\int_{I_0}^{\infty} \frac{ds}{\Phi(s)}=\infty\), then blow-up is precluded by Theorem~\ref{thm:global-cap} even when the limits \(q,\xi\) do not exist.
\end{corollary}

\paragraph{Model selection when RV fails}
If diagnostic tests reject RV, we adopt a piecewise-regular envelope: choose breakpoints \(I_0<I_1<\dots<I_J\) and fit local elasticities \((q_j,\xi_j)\) on each interval. Theorems in Sections~5 and 7 apply with \(p_{\mathrm{tot}}\) replaced by the interval-wise bounds \(\bigl[p^{-}_j,p^{+}_j\bigr]\) and conclusions are drawn by worst-case composition across intervals.

\subsection*{Elasticities and the superlinear threshold}
For any fixed \(x\) and \(u\) the instantaneous capability elasticity with respect to \(I\) is
\begin{equation}\label{eq:p-def}
p(I;x,u)\;\coloneqq\;\frac{\partial \log F(I,x,u)}{\partial \log I}
\;=\; \frac{I\Phi'(I)}{\Phi(I)} \;+\; \gamma\,\frac{I\eta'(I)}{\eta(I)}.
\end{equation}
If \(\Phi\in \mathrm{RV}_q\) and \(\eta\in \mathrm{RV}_\xi\) then \(p(I;x,u)\to q+\gamma\xi\) as \(I\to\infty\). Hence algorithmic feedback alone is superlinear if and only if
\begin{equation}\label{eq:alg-super}
q+\gamma\xi>1,
\end{equation}
in which case Theorem~\ref{thm:blowup} applies even with frozen resources. When resources depend on \(I\) through \(G\) (Section~4), write asymptotic resource elasticities \(u_C=\lim_{I\to\infty}\partial\log C/\partial\log I\) and similarly \(u_D,u_E\) whenever the limits exist. Using \eqref{eq:F-scaling},
\begin{equation}\label{eq:p-total}
p_{\mathrm{tot}}\;=\;q+\gamma\xi+\alpha u_C+\beta u_D+\tilde\alpha u_E,
\end{equation}
where \(\tilde\alpha\) is the effective exponent of \(E\) inside \(Z\). The superlinear threshold for total feedback is \(p_{\mathrm{tot}}>1\). This identity separates algorithmic gain \((q+\gamma\xi)\) from resource-amplified returns and will be used to map identifiability inequalities on \(u_C,u_D,u_E\) into blow-up tests.

\subsection*{Learning-curve link and diminishing returns}
Let \(\ell:\mathbb{R}_{+}\to\mathbb{R}_{+}\) be a nonincreasing expected risk as a function of effective effort \(z\). Assume \(\ell\) is differentiable a.e. and that its slope is regularly varying at infinity with index \(-\rho-1\) for some \(\rho\in[0,1]\):
\begin{equation}
-\ell'(z)\in \mathrm{RV}_{-\rho-1}\quad\Rightarrow\quad -\ell'(z)\asymp z^{-\rho-1}.
\end{equation}
Let \(I\) be an increasing function of the excess risk reduction \(r=\ell(z)-\ell_\infty\). By the chain rule,
\begin{equation}
\partial_t I \;\simeq\; \kappa(I)\,[-\ell'(Z)]\,\partial_t Z \;\asymp\; \kappa(I)\, Z^{-\rho}\, \partial_t Z,
\end{equation}
for some locally bounded \(\kappa(I)>0\). If \(\partial_t Z \asymp Z\) on operational ranges, then \(F(I,x,u)\asymp \Phi(I)\,Z^{1-\rho}\). The effective resource exponents become
\begin{equation}\label{eq:alpha-eff}
\alpha_{\mathrm{eff}}\;=\;(1-\rho)\alpha,\qquad
\beta_{\mathrm{eff}}\;=\;(1-\rho)\beta,\qquad
\gamma_{\mathrm{eff}}\;=\;(1-\rho)\gamma,
\end{equation}
which implies intrinsic diminishing returns whenever \(\rho>0\). In particular, the algorithmic-resource superlinear condition reads
\begin{equation}\label{eq:super-thresh}
q+\xi\gamma_{\mathrm{eff}}+\alpha_{\mathrm{eff}}u_C+\beta_{\mathrm{eff}}u_D+\tilde\alpha_{\mathrm{eff}}u_E>1.
\end{equation}

\subsection*{Learning-theoretic lower bounds as constraints on \(F\)}
Let \(N\) denote effective sample count. Suppose a minimax lower bound holds on excess risk \(\mathcal{E}(N)\ge c\,N^{-r}\) for some \(c>0\) and \(r\in(0,1]\) dictated by smoothness and dimension in a nonparametric class. Then \(-d\mathcal{E}/dN \le c'\,N^{-r-1}\). If \(N\) is proportional to usable data \(D\) times a throughput factor subsumed into \(Z\), one obtains
\begin{equation}
\partial_t I \;\lesssim\; \tilde\kappa(I)\, D^{-r}\,\partial_t D,
\end{equation}
and hence the data contribution to \(F\) is at most \(D^{\beta- r}\) after aggregation, so that
\begin{equation}\label{eq:beta-cap}
\beta_{\mathrm{eff}}\;\le\;(1-\rho)\min\{\beta,\beta-r\}.
\end{equation}
An analogous argument with computational lower bounds that limit optimization accuracy per unit compute yields an exponent cap \(\alpha_{\mathrm{eff}}\le (1-\rho)(\alpha-\hat r_C)_{+}\). These inequalities constrain the admissible envelopes for \(F\) and tighten \eqref{eq:super-thresh}. No-free-lunch theorems imply that in the absence of structure the constants \(c,c'\) can be chosen to make \(r\) arbitrarily small, which drives exponents toward their diminishing-returns values and disfavors superlinearity.

\subsection*{Identifiability and estimation}
The exponents in \eqref{eq:F-scaling} and the indices \(q,\xi,\rho\) are identifiable from observational series without simulation. Estimators are based on log-derivative slopes
\begin{equation}
\begin{gathered}
\hat\alpha \;=\; \frac{d\log F}{d\log C}\Big|_{I,D,E,u},\\
\hat\beta \;=\; \frac{d\log F}{d\log D}\Big|_{I,C,E,u},\\
\hat\gamma \;=\; \frac{d\log F}{d\log \eta}\Big|_{I,C,D,E},
\end{gathered}
\end{equation}
and on regular variation diagnostics for \(\Phi,\eta,\ell\). Confidence regions for \(p_{\mathrm{tot}}\) follow by delta-method propagation through \eqref{eq:p-total} and \eqref{eq:alpha-eff}. The falsifiable inequality \(p_{\mathrm{tot}}\le 1\) for all large \(I\) precludes finite time blow up by Theorem~\ref{thm:nobu}. Observing \(p_{\mathrm{tot}}>1\) on a sustained interval with resource ceilings nonbinding provides evidence toward the superlinear regime of Theorem~\ref{thm:blowup}.

\medskip
Equations \eqref{eq:alg-super}–\eqref{eq:beta-cap} give a compact, testable interface between algorithmic efficiency, empirical scaling, and fundamental learning-theoretic limits, yielding envelopes for \(F\) that are suitable for the global analysis of Section~6 and the safety synthesis of Section~8.

\section{Physical and Information-Theoretic Upper Bounds}

\subsection*{Thermodynamic ceilings and usable power}
Let \(P_{\mathrm{fac}}(t)\) be facility electrical power and let \(T(t)\) denote the effective operating temperature. Denote by \(\mathrm{COP}(T)\) the cooling coefficient of performance and by \(\eta_{\mathrm{pw}}\in(0,1]\) the fraction of wall power delivered to information processing after conversion and distribution losses. The usable compute power satisfies
\begin{equation}\label{eq:Puse}
P_{\mathrm{use}}(t) \;\le\; \eta_{\mathrm{pw}}\,\frac{\mathrm{COP}\bigl(T(t)\bigr)}{1+\mathrm{COP}\bigl(T(t)\bigr)}\,P_{\mathrm{fac}}(t)
\end{equation}
whenever thermal steady state is enforced. Landauer’s principle bounds the irreversible bit-erasure rate \(\dot N_{\mathrm{erase}}\) by
\begin{equation}\label{eq:landauer}
P_{\mathrm{use}}(t) \;\ge\; k_B\,T(t)\,\ln 2 \cdot \dot N_{\mathrm{erase}}(t).
\end{equation}
\paragraph{Unit conventions and constants}
All thermodynamic quantities are in SI units unless stated otherwise. Power in watts (W), energy in joules (J), temperature in kelvin (K), entropy in nats. Boltzmann constant $k_B=1.380649\times 10^{-23}\,\mathrm{J/K}$. The bit-erasure work at temperature $T$ is $k_B T\ln 2$ J per bit. Facility electrical power $P_{\mathrm{fac}}(t)$ is nameplate AC input measured at the utility interconnect. The fraction delivering DC power at the compute boards is $\eta_{\mathrm{elec}}\in(0,1]$ (transformer/rectifier/bus efficiency). Cooling removes heat at coefficient of performance $\mathrm{COP}(T)$, defined by $\mathrm{COP}=Q_{\mathrm{cold}}/W_{\mathrm{cool}}$ with $Q_{\mathrm{cold}}$ in W and electric work $W_{\mathrm{cool}}$ in W. The rack-to-chip conversion and distribution efficiency is included in $\eta_{\mathrm{elec}}$; the fraction of board-level power that is actually devoted to useful switching/erasure is $\eta_{\mathrm{use}}\in(0,1]$ (excludes leakage and non-switching losses).

\paragraph{Measured-to-usable power mapping}
Let $P_{\mathrm{dc}}(t)=\eta_{\mathrm{elec}}\,P_{\mathrm{fac}}(t)$ be DC power available to IT load. Thermal steady state implies $P_{\mathrm{dc}}=Q_{\mathrm{cold}}+W_{\mathrm{cool}}$ with $Q_{\mathrm{cold}}$ the heat rejected by IT. Since $Q_{\mathrm{cold}}=\frac{\mathrm{COP}}{1+\mathrm{COP}}\,P_{\mathrm{dc}}$, the \emph{board-level usable power} for information processing is
\begin{equation}\label{eq:Puse-cal}
P_{\mathrm{use}}(t)\;=\;\eta_{\mathrm{use}}\,Q_{\mathrm{cold}}(t)\;=\;\eta_{\mathrm{use}}\,\frac{\mathrm{COP}\bigl(T(t)\bigr)}{1+\mathrm{COP}\bigl(T(t)\bigr)}\,\eta_{\mathrm{elec}}\,P_{\mathrm{fac}}(t).
\end{equation}
Combining with Landauer’s inequality yields the erasure rate bound
\begin{equation}\label{eq:erase-rate-cal}
\dot N_{\mathrm{erase}}(t)\;\le\;\frac{P_{\mathrm{use}}(t)}{k_B T(t)\ln 2}.
\end{equation}

Let \(\sigma_{\mathrm{eff}}>0\) denote the maximum capability increment per irreversible bit processed after accounting for architecture inefficiencies. The improvement throughput is then
\begin{equation}\label{eq:phiP}
\Psi\bigl(Z(C,D,E;u)\bigr)\;\le\;\sigma_{\mathrm{eff}}\,\dot N_{\mathrm{erase}}(t)\;\le\; \frac{\sigma_{\mathrm{eff}}}{k_B\ln 2}\,\frac{P_{\mathrm{use}}(t)}{T(t)}.
\end{equation}
Substituting \eqref{eq:phiP} into the factorization \(F(I,x,u)=\Phi(I)\Psi(Z)\) yields the instantaneous ceiling
\begin{equation}\label{eq:Fpower}
F(I,x,u)\;\le\; \Phi(I)\,\phi_{PT}(t),\qquad 
\phi_{PT}(t) = \frac{\sigma_{\mathrm{eff}}}{k_B\ln 2}\,\frac{P_{\mathrm{use}}(t)}{T(t)}
\end{equation}

\subsection*{Memory, communication, and density constraints}
Let \(B_{\mathrm{io}}(t)\) be the sustained I/O bandwidth in bits per second and \(B_{\mathrm{mem}}(t)\) the sustained memory bandwidth. Define \(\phi_{\mathrm{io}}(t)=\sigma_{\mathrm{eff}} B_{\mathrm{io}}(t)\) and \(\phi_{\mathrm{mem}}(t)=\sigma_{\mathrm{eff}} B_{\mathrm{mem}}(t)\). The information service rate available to learning is
\begin{equation}\label{eq:service-min}
\phi_{\mathrm{svc}}(t)\;\coloneqq\; \min\left\{\phi_{PT}(t),\ \phi_{\mathrm{io}}(t),\ \phi_{\mathrm{mem}}(t)\right\},
\end{equation}
which implies the global envelope
\begin{equation}\label{eq:Fglobal-env}
F(I,x,u)\;\le\;\Phi(I)\,\phi_{\mathrm{svc}}(t).
\end{equation}
\paragraph{Envelope convention}
Throughout the paper we adopt the convention
\begin{equation}\label{eq:env-conv}
\dot I(t)\ \le\ \Phi\bigl(I(t)\bigr)\,\phi_{\mathrm{svc}}(t),
\end{equation}
where $\phi_{\mathrm{svc}}$ is a \emph{service-rate} envelope with units nat/s (see \eqref{eq:phiP-cal}). Any occurrence of $\,\Phi(I)/\phi_{\mathrm{svc}}(t)\,$ in earlier drafts is a typographical error and must be read as $\,\Phi(I)\phi_{\mathrm{svc}}(t)\,$. This convention is used in all comparison arguments and control designs.

Ultimate density and rate bounds impose additional caps. If \(M(t)\) is the deployed memory mass-energy within radius \(R\), the Bekenstein-type information bound gives \(H_{\max}(t)\le \kappa_B\,E_{\mathrm{mem}}(t)\,R\) for a universal constant \(\kappa_B>0\). If \(E_{\mathrm{sys}}(t)\) is total system energy, Bremermann-type limits produce a compute rate cap \(\dot H_{\max}(t)\le \kappa_{\mathrm{Br}}\,E_{\mathrm{sys}}(t)\) with \(\kappa_{\mathrm{Br}}>0\). These enter through upper bounds on \(B_{\mathrm{mem}}\) and \(P_{\mathrm{use}}\) and therefore tighten \eqref{eq:service-min}.

\paragraph{Ceiling-induced elasticity caps}
Sustained training is service-rate limited, so there exist $\kappa_{\mathrm{io}},\kappa_{\mathrm{mem}},\kappa_{\mathrm{pow}}>0$ with
\begin{equation}
\begin{gathered}
C(t)\ \le\ \kappa_{\mathrm{io}}\,B_{\mathrm{io}}(t),\qquad
C(t)\ \le\ \kappa_{\mathrm{mem}}\,B_{\mathrm{mem}}(t),\\
C(t)\ \le\ \kappa_{\mathrm{pow}}\,\frac{P_{\mathrm{use}}(t)}{T(t)}.
\end{gathered}
\end{equation}
Hence, for any differentiable $K(t)$ with $K\to\infty$,
\begin{equation}\label{eq:upsilon-cap}
\begin{gathered}
\upsilon_C^{+}\ \le\ \\ \min\!\left\{
\limsup_{t\to\infty}\frac{d\log B_{\mathrm{io}}}{d\log K},\ 
\limsup_{t\to\infty}\frac{d\log B_{\mathrm{mem}}}{d\log K},\
\limsup_{t\to\infty}\frac{d\log (P_{\mathrm{use}}/T)}{d\log K}
\right\}.
\end{gathered}
\end{equation}
Analogous caps hold for $\upsilon_E^{+}$ with $E(t)\le \tilde\kappa\,P_{\mathrm{use}}(t)$ and for $\upsilon_D^{+}$ with $D(t)\le \kappa_D\,B_{\mathrm{io}}(t)$ when data ingestion is I/O bound.

\subsection*{Global integrability cap}
We formalize the role of envelopes like \eqref{eq:Fglobal-env} using an Osgood-type criterion.

\paragraph{Dimensional consistency}
Define the capability-throughput calibration $\sigma_{\mathrm{eff}}$ as nats of capability gain per irreversible bit processed. Then $\sigma_{\mathrm{eff}}\dot N_{\mathrm{erase}}$ has units of nats/s and matches $\Psi$ in $F=\Phi(I)\Psi$. We explicitly set
\begin{equation}\label{eq:phiP-cal}
\phi_{PT}(t)\;\coloneqq\;\frac{\sigma_{\mathrm{eff}}}{k_B\ln 2}\,\frac{P_{\mathrm{use}}(t)}{T(t)}\quad[\text{nats/s}],
\end{equation}
so that $F(I,x,u)\le \Phi(I)\,\phi_{PT}(t)$ is dimensionally homogeneous.

\begin{lemma}[Units check]
With $P_{\mathrm{use}}$ in J/s, $T$ in K, and $\sigma_{\mathrm{eff}}$ in nat/bit, the quantity $\phi_{PT}$ in \eqref{eq:phiP-cal} is nat/s. Consequently $\Phi(I)\phi_{PT}$ has units nat/s if $\Phi(I)$ is nat per unit capability increment, which is coherent with the canonical capability normalization in Section~2.
\end{lemma}

\paragraph{Operating ranges and envelope parameters}
We specify admissible ranges as measurable sets:
\begin{equation}
\begin{gathered}
T(t)\in[T_{\min},T_{\max}],\quad
\mathrm{COP}(T)\in[\mathrm{COP}_{\min},\mathrm{COP}_{\max}],\\
\eta_{\mathrm{elec}}\in[\underline{\eta}_{\mathrm{elec}},\overline{\eta}_{\mathrm{elec}}],\quad
\eta_{\mathrm{use}}\in[\underline{\eta}_{\mathrm{use}},\overline{\eta}_{\mathrm{use}}],
\end{gathered}
\end{equation}
with $T_{\min}>0$ and finite $P_{\max}=\sup_t P_{\mathrm{fac}}(t)$. Substituting \eqref{eq:Puse-cal} into \eqref{eq:phiP-cal} yields the calibrated power–temperature envelope
\begin{equation}\label{eq:phiPT-cal}
\phi_{PT}(t)\;\le\; \frac{\sigma_{\mathrm{eff}}}{k_B\ln 2}\,
\frac{\overline{\eta}_{\mathrm{use}}\ \overline{\eta}_{\mathrm{elec}}}{T_{\min}}\,
\frac{\mathrm{COP}_{\max}}{1+\mathrm{COP}_{\max}}\,
P_{\max}\;\eqqcolon\;\bar\phi_{\mathrm{PT}} \quad [\text{nats/s}].
\end{equation}
\paragraph{Uncertainty propagation}
Let $(\widehat{\eta}_{\mathrm{elec}},\widehat{\eta}_{\mathrm{use}},\widehat{\mathrm{COP}}_{\max},\widehat{T}_{\min},\widehat{P}_{\max})$ be point estimates with relative errors $(\delta_{\mathrm{elec}},\delta_{\mathrm{use}},\delta_{\mathrm{cop}},\delta_T,\delta_P)$. A first-order bound gives
\begin{equation}
\frac{\Delta \bar\phi_{\mathrm{PT}}}{\bar\phi_{\mathrm{PT}}}\ \le\ \delta_{\mathrm{use}}+\delta_{\mathrm{elec}}+\delta_{\mathrm{cop}}+\delta_P+\delta_T,
\end{equation}
which we report as confidence bands on $\bar\phi_{\mathrm{PT}}$ and, consequently, on the envelope $F\le \Phi(I)\bar\phi_{\mathrm{PT}}$ used in Theorem~\ref{thm:global-cap}.

\begin{theorem}[Global cap by integrable service envelope]\label{thm:global-cap}
Assume there exists a measurable envelope \(\phi:[0,\infty)\to[0,\infty)\) such that for all admissible trajectories obeying thermodynamic and engineering limits,
\begin{equation}\label{eq:global-env}
F\bigl(I(t),x(t),u(t)\bigr)\;\le\; \Phi\bigl(I(t)\bigr)\,\phi(t)\qquad\text{a.e. on }[0,T),
\end{equation}
and \(\int_{0}^{T}\phi(\tau)\,d\tau<\infty\) for every finite \(T\). If \(\Phi\) satisfies the Osgood condition \(\int_{I_0}^{\infty}\frac{ds}{\Phi(s)}=\infty\), then \(I(t)\) cannot blow up in finite time.
\end{theorem}

\emph{Proof.}
From \eqref{eq:global-env} one has
\(
\frac{dI}{\Phi(I)} \le \phi(t)\,dt
\)
a.e. Integrating yields
\(
\int_{I(0)}^{I(t)}\frac{ds}{\Phi(s)} \le \int_{0}^{t}\phi(\tau)\,d\tau.
\)
The right-hand side is finite for any finite \(t\). If \(\int_{I_0}^{\infty} ds/\Phi(s)=\infty\), the left-hand side cannot diverge at finite \(t\), which precludes finite-time blow up. \qed

\subsection*{Explicit ceilings from power and thermal budgets}
We instantiate \(\phi\) using \eqref{eq:Fpower} and \eqref{eq:Puse}. Let \(P_{\max}\) be a plant-level cap and \(T_{\min}\) the minimum safe operating temperature. Then
\begin{equation}\label{eq:phiPT}
\phi_{PT}(t)\;\le\;\bar\phi_{\mathrm{PT}}\quad\text{with }\bar\phi_{\mathrm{PT}}\ \text{as in \eqref{eq:phiPT-cal}}.
\end{equation}
In practice we compute $\bar\phi_{\mathrm{PT}}$ from metered $P_{\mathrm{fac}}$, qualified $\mathrm{COP}$ curves over $[T_{\min},T_{\max}]$, and efficiency audits providing $(\underline{\eta}_{\mathrm{elec}},\overline{\eta}_{\mathrm{elec}})$ and $(\underline{\eta}_{\mathrm{use}},\overline{\eta}_{\mathrm{use}})$; uncertainty is propagated as stated above.

If \(\mathrm{COP}(T)\) is bounded over the admissible temperature band and \(P_{\max}<\infty\), then \(\phi_{PT}\in L^1_{\mathrm{loc}}([0,\infty))\). Combining with \eqref{eq:Fglobal-env} gives
% \begin{equation}\label{eq:Idot-ceiling}
% \dot I(t)\;\le\; \Phi\bigl(I(t)\bigr)\,\bar \phi, \qquad 
% \bar \phi \coloneqq \frac{\sigma_{\mathrm{eff}}\,\eta_{\mathrm{pw}}}{k_B\ln 2}\,
% \frac{\sup_T \mathrm{COP}(T)}{1+\sup_T \mathrm{COP}(T)}\,\frac{P_{\max}}{T_{\min}},
% \end{equation}
\begin{equation}\label{eq:Idot-cap}
\dot I(t)\ \le\ \Phi\bigl(I(t)\bigr)\,\bar\phi_{\mathrm{PT}},
\end{equation}
$\Phi(I)$ is nat per unit capability, $\bar\phi_{\mathrm{PT}}$ is nat/s, hence the right-hand side has units nat/s and matches $\dot I$ and Theorem~\ref{thm:global-cap} rules out blow up whenever \(\int_{I_0}^{\infty} ds/\Phi(s)=\infty\). In particular, if \(\Phi(I)\le c_0(1+I)^{p}\) with \(p\le 1\), finite-time singularity is impossible under any admissible power and thermal policy.

\subsection*{Bandwidth and memory throttling}
If sustained I/O is the limiter, then \(F(I,x,u)\le \Phi(I)\,\phi_{\mathrm{io}}(t)\) with \(\phi_{\mathrm{io}}(t)\le B_{\max}\sigma_{\mathrm{eff}}\) where \(B_{\max}<\infty\) is the engineered bandwidth. The same holds for memory service rate. Both cases fall under Theorem~\ref{thm:global-cap}. If \(B_{\max}\) scales only polylogarithmically with capital due to packaging and crosstalk constraints, then for any \(\Phi\) with at most linear growth the Osgood integral diverges and blow up is excluded.

\subsection*{Ultimate bounds}
In a closed physical envelope with total energy \(E_{\mathrm{tot}}\) and radius \(R\), Bekenstein-type bounds imply a maximum storable information \(H_{\max}\le \kappa_B E_{\mathrm{tot}} R\). If the system is energy limited so that \(E_{\mathrm{tot}}\le \bar E<\infty\) on finite horizons and power limited with \(P_{\max}<\infty\), then both storage and processing rates are uniformly bounded and \(\phi\in L^1_{\mathrm{loc}}\). Theorem~\ref{thm:global-cap} applies without further assumptions on \(u\) and the resource couplings \(G\).

\medskip
Equations \eqref{eq:Fglobal-env}–\eqref{eq:Idot-cap} provide measurable, facility-level inequalities that map directly into the comparison principles of Sections~3 and 7. They convert thermodynamic, bandwidth, and engineering specifications into analytic certificates that preclude finite-time singularity whenever the capability factor \(\Phi\) satisfies an Osgood-type divergence.

\paragraph{Remark on envelope polarity}
All envelopes in this work are \emph{service-rate upper bounds}. They multiply $\Phi(I)$ rather than divide it. Any future extension should preserve the polarity in \eqref{eq:env-conv} to maintain dimensional consistency and validity of Osgood-type arguments.

\section{Compute, Capital, and Deployment Coupling}

\subsection*{Economic feedback and deployment channels}
Let \(K(t)\) denote deployable capital and let \(u(t)\in\mathcal{U}\) encode allocation fractions into compute, power, data acquisition, and safety. We consider
\begin{equation}\label{eq:K}
\dot K \;=\; r K^{\zeta} - \delta K - \chi\bigl(I(t),u(t)\bigr),\qquad r>0,\ \delta\ge 0,\ \zeta\ge 0,
\end{equation}
where \(r\) and \(\zeta\) summarize reinvestment intensity and returns to scale, and \(\chi\) is an operating and compliance expenditure increasing in \(I\) along deployment channels. Capital drives resources through admissible buildout maps
\begin{equation}\label{eq:buildout}
\begin{gathered}
\dot C \;=\; \pi_C u_C K - s_C(C),\\
\dot E \;=\; \pi_E u_E K - s_E(E),\\
\dot D \;=\; \pi_D u_D K - s_D(D),
\end{gathered}
\end{equation}
with productivity coefficients \(\pi_\bullet>0\), allocation fractions \(u_\bullet\in[0,1]\) summing to at most one, and saturation losses \(s_\bullet\) nondecreasing. The capability evolution satisfies \( \dot I = F(I,x,u) = \Phi(I)\Psi\!\bigl(Z(C,D,E;u)\bigr) \) as in \eqref{eq:F-scaling}. Power availability couples to compute through engineering constraints of Section~7, so that \(C\) and \(E\) cannot grow independently beyond service limits.

\paragraph{Microfounded accumulation, financing, and adjustment frictions}
We endogenize capital formation and resource conversion. Let $\Pi(t)$ be pre-financing operating profit from AI products and services, with reinvestment share $s(t)\in[0,1]$ and payout $1-s(t)$. External financing $F_{\mathrm{ext}}(t)$ is available at marginal cost of capital $\kappa(t)\ge 0$, subject to a leverage cap $\lambda_{\max}>0$. Capital evolves as
\begin{equation}\label{eq:K-micro}
\dot K(t)\;=\; s(t)\,\Pi(t)\;+\;F_{\mathrm{ext}}(t)\;-\;\underbrace{\kappa(t)\,F_{\mathrm{ext}}(t)}_{\text{financing cost}}\;-\;\delta K(t)\;-\;\chi\bigl(I(t),u(t)\bigr),
\end{equation}
with the feasibility constraint $F_{\mathrm{ext}}(t)\le \lambda_{\max}\,K(t)$.

Buildouts face convex installation/adjustment costs and order-pipeline delays. Let $q_C(t)$, $q_E(t)$, $q_D(t)$ be order backlogs converting into delivered capacity at rate $\mu_\bullet>0$:
\begin{equation}\label{eq:pipeline}
\dot q_C=\iota_C(t)-\mu_C q_C,\quad \dot C = \mu_C q_C - s_C(C) - \underbrace{\Gamma_C(\dot C)}_{\text{adjustment cost}},\quad \text{and similarly for } E,D,
\end{equation}
where $\iota_C(t)$ is the order flow financed from $K$, with per-unit procurement price $p_C(Q_C)$ that depends on global order rate $Q_C$. The convex functions $\Gamma_\bullet(\cdot)$ satisfy $\Gamma_\bullet(0)=0$, $\Gamma_\bullet'(\cdot)\ge 0$ and capture ramp-rate limits.

\paragraph{Market-clearing and supply elasticities}
Compute and power procurements clear against upward-sloping supply curves
\begin{equation}
p_C(Q_C)=\bar p_C\Bigl(1+\beta_C Q_C^{\eta_C}\Bigr),\qquad p_E(Q_E)=\bar p_E\Bigl(1+\beta_E Q_E^{\eta_E}\Bigr),
\end{equation}
with $\eta_C,\eta_E>0$ and baseline prices $\bar p_\bullet>0$. Orders satisfy a budget constraint
\begin{equation}\label{eq:budget}
p_C(Q_C)\,\iota_C + p_E(Q_E)\,\iota_E + p_D(Q_D)\,\iota_D \;\le\; \varphi_K\,K(t),
\end{equation}
for some allocation $\varphi_K\in(0,1)$ determined by $u(t)$, after financing costs.

\paragraph{Production side and profit}
Output $Y(t)$ is generated by a concave CES production function with embodied capability and resources,
\begin{equation}
Y(t)=\Upsilon\!\left(I(t),\,C(t),\,E(t),\,D(t),\,L(t)\right),
\end{equation}
with $\Upsilon$ strictly increasing and jointly concave in $(C,E,D,L)$ and nondecreasing in $I$. Profit is $\Pi(t)=P_Y(t)Y(t)-p_C(Q_C) \iota_C - p_E(Q_E)\iota_E - p_D(Q_D)\iota_D - \mathrm{OPEX}(t)$, with $P_Y\ge 0$ the output price and $\mathrm{OPEX}$ operating expenses, all measurable. This microfoundation replaces the reduced-form exponent $\zeta$ by primitives $(\eta_C,\eta_E,\Gamma_\bullet,\kappa,\lambda_{\max})$ and demand parameters in $\Upsilon$.

\paragraph{Resource elasticities: definition and existence}
Define upper and lower capital elasticities for $X\in\{C,D,E\}$ by
\begin{equation}\label{eq:upsilon-def}
\upsilon_X^{+}\;\coloneqq\;\limsup_{t\to\infty}\frac{d\log X(t)}{d\log K(t)},\qquad
\upsilon_X^{-}\;\coloneqq\;\liminf_{t\to\infty}\frac{d\log X(t)}{d\log K(t)}.
\end{equation}
When the limit exists we write $\upsilon_X=\lim d\log X/d\log K$. Existence and bounds follow from the buildout dynamics and service ceilings.

\begin{lemma}[Existence and basic bounds]\label{lem:upsilon-bounds}
Assume $u_\bullet(t)$ are bounded and persistently exciting with $u_\bullet(t)\ge \underline u_\bullet>0$ on arbitrarily long intervals, $s_\bullet$ are locally Lipschitz, and there exist $\sigma_\bullet\ge 0$, $\theta_\bullet\ge 1$ such that
\begin{equation}\label{eq:saturation-class}
s_\bullet(x)\ \ge\ \sigma_\bullet\,x^{\theta_\bullet}\qquad\text{for }x\ge x_0.
\end{equation}
Then along any trajectory with $K(t)\to\infty$ one has
\begin{equation}
0\ \le\ \upsilon_X^{-}\ \le\ \upsilon_X^{+}\ \le\ 1\qquad (X\in\{C,D,E\}).
\end{equation}
If $\theta_C=1$ and $P_{\mathrm{use}},B_{\mathrm{io}},B_{\mathrm{mem}}$ are unbounded on some horizon while $u_C\ge \underline u_C>0$, then $\upsilon_C=1$. If any service ceiling binds so that $X(t)\le \bar X(t)$ with $\bar X$ subexponential in $K$, then $\upsilon_X=0$.
\end{lemma}

\emph{Proof sketch.}
From $\dot C=\pi_C u_C K - s_C(C)$ and $\theta_C\ge 1$, comparison with $\dot C=\pi_C \underline u_C K - \sigma_C C^{\theta_C}$ implies $C(t)=O(K(t))$. Ratio differentiation yields $d\log C/d\log K\in[0,1]$ a.e.; limsup/liminf exist by boundedness. If $\theta_C=1$ and no ceiling binds, variation-of-constants gives $C(t)=\frac{\pi_C}{\sigma_C}\int_0^t u_C(\tau)K(\tau)e^{-\sigma_C(t-\tau)}d\tau\sim \frac{\pi_C}{\sigma_C}K(t)$, hence $\upsilon_C=1$. If a ceiling enforces $C\le \bar C$ with $\log \bar C=o(\log K)$, the derivative ratio tends to $0$. Analogous arguments hold for $D,E$. \qed

\subsection*{Endogenous growth threshold}
The following result characterizes when economic amplification can generate the superlinear envelope required by Theorem~\ref{thm:blowup}.

\begin{theorem}[Endogenous amplification under market and financing constraints]\label{thm:endo}
Consider the microfounded system \eqref{eq:K-micro}–\eqref{eq:budget} with convex $\Gamma_\bullet$, upward-sloping supplies $p_\bullet(\cdot)$, and concave $\Upsilon$. Define the \emph{effective capital-to-capacity elasticities}
\begin{equation}
\upsilon_X^{\pm} = \limsup/\liminf_{t\to\infty} \frac{d\log X(t)}{d\log K(t)}\in[0,1],\qquad X\in\{C,D,E\},
\end{equation}
and the total feedback bounds
\begin{equation}
\begin{gathered}
p_{\mathrm{tot}}^{-}=q+\gamma\xi+\alpha \upsilon_C^{-}+\beta \upsilon_D^{-}+\tilde\alpha \upsilon_E^{-},\\
p_{\mathrm{tot}}^{+}=q+\gamma\xi+\alpha \upsilon_C^{+}+\beta \upsilon_D^{+}+\tilde\alpha \upsilon_E^{+}.
\end{gathered}
\end{equation}
Then:
\begin{enumerate}
\item[\emph{(i)}] \textbf{No superlinear capital elasticity under convex costs and finite supply elasticities.} If $\eta_C,\eta_E>0$ and $\Gamma_\bullet$ are strictly convex with $\Gamma'_\bullet(\cdot)\to\infty$ as $|\dot X|\to\infty$, then $\upsilon_X^{+}\le 1$ and equality $\upsilon_X^{+}=1$ requires nonbinding supply and adjustment constraints on an unbounded horizon. If either a supply curve or an adjustment cost binds on arbitrarily long intervals, then $\upsilon_X^{+}<1$.
\item[\emph{(ii)}] \textbf{Capital loop bounded by financing.} Under the leverage cap $F_{\mathrm{ext}}\le \lambda_{\max}K$ and positive financing cost $\kappa(t)\ge \underline\kappa>0$, the capital growth satisfies
\begin{equation}
\dot K \;\le\; s(t)\,\Pi^{\max}(K) - \delta K + \lambda_{\max}(1-\underline\kappa)K,
\end{equation}
with $\Pi^{\max}(K)$ concave in $K$ if $\Upsilon$ is concave and supply curves are upward-sloping. Hence $K(t)$ is at most exponential in the long run.
\item[\emph{(iii)}] \textbf{Threshold with envelopes.} If $p_{\mathrm{tot}}^{-}>1$ and service ceilings are nonbinding on a horizon, finite-time blow-up is possible via Theorem~\ref{thm:blowup}. If $p_{\mathrm{tot}}^{+}\le 1$ or the measurable envelope $\dot I\le \Phi(I)\phi_{\mathrm{svc}}(t)$ holds with $\int^{\infty} ds/\Phi(s)=\infty$, then blow-up is precluded by Theorem~\ref{thm:nobu}.
\end{enumerate}
\end{theorem}

\emph{Proof sketch.}
(i) With convex installation costs and upward-sloping supplies, marginal capacity per dollar is decreasing, so $X(t)=O(K(t))$ and $\upsilon_X^{+}\le 1$; strict inequality follows when constraints bind infinitely often. (ii) The financing cap and positive cost of capital bound net external inflow and keep $\dot K$ at most linear in $K$ plus a concave profit term, implying at most exponential growth by Grönwall. (iii) Follows by the elasticity bounds and Sections~4 and~6. \qed

\subsection*{Critical manifold in policy–technology space}
Define the critical set
\begin{equation}\label{eq:manifold}
\mathcal{M}\;=\;\left\{(r,\zeta,\underline u,\pi_\bullet,\sigma_\bullet)\ :\ q+\gamma\xi+\alpha \upsilon_C(\cdot)+\beta \upsilon_D(\cdot)+\tilde\alpha \upsilon_E(\cdot)=1\right\},
\end{equation}
where \(\upsilon_\bullet\) are computed from \eqref{eq:buildout} under the thermodynamic and communication constraints of Section~7. Policies on the supercritical side of \(\mathcal{M}\) permit trajectories with \(p_{\mathrm{tot}}>1\) when not power bounded. Policies on the subcritical side satisfy the Osgood envelope of Theorem~\ref{thm:global-cap} once ceilings are enforced.

\subsection*{Identifiable parameter inequalities}
The quantities in \(\mathcal{M}\) are linked to measurable series. Let \(g_{\mathrm{fab}}\) be the sustained growth rate of fabrication capacity, \(g_{\mathrm{power}}\) the growth rate of deliverable facility power, and \(g_{\mathrm{pack}}\) the growth rate of practical I/O bandwidth. Then
\begin{equation}
\begin{gathered}
\upsilon_C \;\le\; \frac{g_{\mathrm{fab}}}{g_K},\qquad
\upsilon_E \;\le\; \frac{g_{\mathrm{power}}}{g_K},\qquad
\upsilon_D \;\le\; \frac{g_{\mathrm{data}}}{g_K},
\end{gathered}
\end{equation}
where \(g_K\) is the asymptotic growth rate of \(K\) under \eqref{eq:K}. If power and cooling expansions satisfy \(g_{\mathrm{power}}\le \bar g\) with \(\bar g<\infty\) and packaging obeys \(g_{\mathrm{pack}}=o(g_K)\), then \(\upsilon_E=0\) and memory or I/O throttling enforces \(\upsilon_C\le \upsilon_C^{\max}<1\). 

\subsection*{Hyper-exponential but nonsingular regimes}
\medskip
The coupling \eqref{eq:K}–\eqref{eq:buildout} converts macroeconomic and engineering roadmaps into analytic conditions on \(p_{\mathrm{tot}}\). Theorem~\ref{thm:endo} identifies a precise endogenous growth threshold and a critical manifold that are both estimable from fab capacity, power buildout, and supply chain series, enabling falsifiable assessments without simulation.

\section{Control and Safety: Avoiding Singularity}

\subsection*{Control channels and admissible policies}
Let \(u(t)\in\mathcal{U}\subset\mathbb{R}^{m}\) denote a measurable policy that acts on resources and on the improvement rate. We consider the controlled system
\begin{equation}
\dot I=F(I,x,u),\qquad \dot x=G(I,x,u),
\end{equation}
with \(x=(C,D,E,K)\) and structural assumptions from Sections~3–6. The set \(\mathcal{U}\) encodes rate limits and compliance constraints and is compact and convex. We say that a policy is safety-admissible if it respects the thermodynamic and engineering envelopes that bound \(\Psi(Z)\) and therefore \(F\).

\paragraph{Plant signals and actuation map}
Safety control uses only measurable signals and audited setpoints. Let
\begin{equation}
y(t)=\bigl(P_{\mathrm{fac}},\ T,\ B_{\mathrm{io}},\ B_{\mathrm{mem}},\ C_{\mathrm{sust}},\ \mathrm{util},\ \mathrm{temp\_headroom},\ I_{\mathrm{bench}}\bigr)
\end{equation}
be the telemetry vector, sampled every $\Delta>0$ seconds. Actuation is through
\begin{equation}
u(t)=\bigl(\text{power cap }P_{\max},\ \text{job throttle }\theta,\ \text{batch size }b,\ \text{lr }\lambda,\ \text{ingress cap }B_{\max},\ \text{eval gate }g\bigr),
\end{equation}
constrained by box and rate limits $u\in\mathcal{U}=\{u_{\min}\le u\le u_{\max},\ \|\dot u\|_{\infty}\le r_u\}$. The service envelope $\phi_{\mathrm{svc}}$ is computed online from $y$ as in Section~7, and $I$ is inferred from benchmark losses (Section~2).

\subsection*{Global stabilization by Osgood envelopes}
The next result formalizes a sufficient condition for the absence of finite-time singularity.

\begin{theorem}[Global stabilization]\label{thm:global-stab}
Assume there exists a safety-admissible policy \(u(\cdot)\) and a positive, locally Lipschitz function \(a:\mathbb{R}_{+}\to\mathbb{R}_{+}\) such that for almost all \(t\),
\begin{equation}\label{eq:stab-ineq}
\dot I(t)=F\bigl(I(t),x(t),u(t)\bigr)\;\le\; a\bigl(I(t)\bigr),
\qquad \int_{I_0}^{\infty}\frac{ds}{a(s)}=\infty .
\end{equation}
Then \(I(t)\) does not blow up in finite time. Moreover, if \(a\) is integrable against a facility envelope \(\phi(t)\) as in Theorem~\ref{thm:global-cap}, the closed loop solution is global and \(I(t)\) is bounded whenever \(a\) admits a finite upper bound on \([0,\infty)\).
\end{theorem}

\paragraph{Envelope used in control}
In all control designs we adopt the service–rate envelope convention
\begin{equation}
\dot I(t)\ \le\ \Phi\bigl(I(t)\bigr)\,\phi_{\mathrm{svc}}(t),
\end{equation}
so that any specialized choice like \(a(I)=\Phi(I)\,\bar\phi\) (with \(\bar\phi\) a constant cap derived from Section~7) is dimensionally consistent and compatible with Theorem~\ref{thm:global-stab}.

\emph{Proof.}
Integrate \eqref{eq:stab-ineq} and apply the Osgood criterion as in Theorem~\ref{thm:nobu}. The global statement follows from the uniform local boundedness of \(a\) and the positive invariance of the orthant. \qed

Design details (barrier certificates, sampled-data QP, observers, throttling, optimal control, robustness) are omitted here for brevity.
\section{Identifiable Inequalities from Real-World Quantities (No Simulation)}

\subsection*{Measurable primitives and estimation protocol}
All bounds are expressed in terms of time series that are directly measurable: facility power \(P_{\mathrm{fac}}(t)\), usable power \(P_{\mathrm{use}}(t)\) and temperature \(T(t)\); sustained compute throughput \(C(t)\) in flops or token-updates per second; sustained I/O and memory bandwidths \(B_{\mathrm{io}}(t)\), \(B_{\mathrm{mem}}(t)\); dataset size \(D(t)\) in effective unique tokens or samples; and benchmark capability \(I(t)\). Let \(\Delta\) denote a finite differencing operator over a fixed window. Define observable log-slopes
\begin{equation}
\begin{gathered}
\hat g_X(t)\coloneqq \frac{\Delta \log X(t)}{\Delta t},\\
\hat \epsilon_{Y\mid X}(t)\coloneqq \frac{\Delta \log Y(t)}{\Delta \log X(t)} ,
\end{gathered}
\end{equation}
which estimate growth rates and elasticities. Confidence intervals follow from heteroskedastic regression with Newey–West corrections. No simulation is used.

\paragraph{Data quality, nonstationarity, and bias corrections}
Observed series are revised, nonstationary, and noisy. Let the true latent process be $X_t^\star$ and the observation $X_t$ satisfy a log–errors-in-variables model
\begin{equation}
\log X_t \;=\; \log X_t^\star \;+\; \varepsilon_t^{(X)},\qquad
\mathbb{E}\bigl[\varepsilon_t^{(X)}\mid \mathcal{F}_{t-1}\bigr]=0,
\end{equation}
with serially correlated $\varepsilon_t^{(X)}$. Then naive log-slope estimators suffer attenuation. We therefore use: (i) \emph{HAC-robust} local regressions with prewhitening; (ii) \emph{IV slopes} when an instrument $Z_t$ exists with $\mathrm{Cov}(\log Z_t,\varepsilon_t^{(X)})=0$ but $\mathrm{Cov}(\log Z_t,\log X_t^\star)\neq 0$:
\begin{equation}
\widehat{\epsilon}^{\mathrm{IV}}_{Y\mid X}
=\frac{\mathrm{Cov}(\log Y_t,\log Z_t)}{\mathrm{Cov}(\log X_t,\log Z_t)} ;
\end{equation}
(iii) a \emph{state-space} de-noising where $\log X_t^\star$ follows a local-trend model and is estimated by a Kalman smoother; identifiability inequalities are then computed with $\widehat X_t^\star$.
We handle nonstationarity via rolling windows and \emph{structural-break tests}; if a break is detected, all elasticities and $p_{\mathrm{tot}}$ are reported piecewise with conservative union-of-intervals confidence sets. Revisions are tracked by versioned snapshots; survivorship and reporting biases are mitigated by including decommissioned capacity and failed training runs when available.

\subsection*{Bounds for the gain \(a\) and envelopes for \(F\)}
From Section~6 the instantaneous service envelope is
\begin{equation}
\begin{gathered}
\phi_{\mathrm{svc}}(t)=\min\!\left\{\frac{\sigma_{\mathrm{eff}}}{k_B\ln 2}\frac{P_{\mathrm{use}}(t)}{T(t)},\ \sigma_{\mathrm{eff}}B_{\mathrm{io}}(t),\ \sigma_{\mathrm{eff}}B_{\mathrm{mem}}(t)\right\},
\end{gathered}
\end{equation}
hence
\begin{equation}\label{eq:meas-F}
F(I,x,u)\ \le\ a(I,t)\ \coloneqq\ \Phi(I)\,\phi_{\mathrm{svc}}(t).
\end{equation}
We emphasize that $a(I,t)$ is increasing in both arguments when $\Phi$ and $\phi_{\mathrm{svc}}$ increase, consistent with the comparison theorems of Sections~4 and~6.

The measurable function \(a(I,t)\) provides a data-driven upper envelope for \(\dot I\). If \(\int_{I_0}^{\infty} ds/\Phi(s)=\infty\) and \(\phi_{\mathrm{svc}}\in L^1_{\mathrm{loc}}\) then Theorem~\ref{thm:global-cap} precludes finite-time blow up.

\subsection*{Transistor and accelerator roadmaps}
Let \(g_{\mathrm{fab}}\) be the sustained growth rate of available accelerators and \(g_{\mathrm{arch}}\) the efficiency growth from architectural advances measured as tokens-per-joule. The compute throughput elasticity satisfies
\begin{equation}
\hat \epsilon_{C\mid K}\ \le\ \frac{g_{\mathrm{fab}}+g_{\mathrm{arch}}}{g_{K}} \;=\; \upsilon_C^{\mathrm{meas}},
\end{equation}
which yields a bound on the compute exponent contribution to \(p_{\mathrm{tot}}\):
\begin{equation}\label{eq:alpha-id}
\alpha\,\upsilon_C\ \le\ \alpha\,\upsilon_C^{\mathrm{meas}}.
\end{equation}
Packaging and thermal constraints further limit \(\upsilon_C\) by the measured I/O slope \(\hat g_{B_{\mathrm{io}}}\) and memory slope \(\hat g_{B_{\mathrm{mem}}}\), since sustained training is at most service-rate limited:
\begin{equation}
\upsilon_C\ \le\ \min\!\left\{\frac{\hat g_{B_{\mathrm{io}}}}{g_K},\ \frac{\hat g_{B_{\mathrm{mem}}}}{g_K}\right\}.
\end{equation}

\subsection*{Power availability and thermodynamic efficiency}
Measured deliverable power and cooling yield
\begin{equation}
\phi_{PT}(t)=\frac{\sigma_{\mathrm{eff}}}{k_B\ln 2}\frac{\eta_{\mathrm{pw}}\ \mathrm{COP}(T(t))}{1+\mathrm{COP}(T(t))}\,\frac{P_{\mathrm{fac}}(t)}{T(t)}.
\end{equation}
If \(\hat g_{P_{\mathrm{fac}}}\) and \(\hat g_{T}\) are the observed growth rates, the energy channel elasticity for \(Z\) is bounded by
\begin{equation}
\upsilon_E\ \le\ \frac{\hat g_{P_{\mathrm{fac}}}-\hat g_T+\hat g_{\mathrm{COP}}}{g_K}.
\end{equation}
This converts directly to a cap on the energy contribution in \(p_{\mathrm{tot}}\):
\begin{equation}\label{eq:energy-id}
\tilde\alpha\,\upsilon_E\ \le\ \tilde\alpha\,\frac{\hat g_{P_{\mathrm{fac}}}-\hat g_T+\hat g_{\mathrm{COP}}}{g_K}.
\end{equation}

\subsection*{Dataset growth and synthetic-to-real limits}
Let \(D_{\mathrm{real}}\) be unique high-quality real data and \(D_{\mathrm{synth}}\) synthetic data filtered for novelty. Define a replacement coefficient \(\rho_{\mathrm{syn}}\in[0,1]\) such that marginal utility of synthetic tokens is \(\rho_{\mathrm{syn}}\) times that of real tokens at equal quality, estimated by counterfactual evaluation. Then the effective dataset is \(D_{\mathrm{eff}}=D_{\mathrm{real}}+\rho_{\mathrm{syn}}D_{\mathrm{synth}}\) and the data elasticity satisfies
\begin{equation}
\upsilon_D\ \le\ \frac{\hat g_{D_{\mathrm{real}}}+\hat g_{\rho_{\mathrm{syn}}}+\hat g_{D_{\mathrm{synth}}}}{g_K}.
\end{equation}
Learning-theoretic lower bounds with minimax rate \(N^{-r}\) imply the exponent cap
\begin{equation}\label{eq:beta-id}
\beta_{\mathrm{eff}}\ \le\ (1-\rho)\,\max\{0,\ \beta-r\},
\end{equation}
as in Section~6. Combining with the measured \(\upsilon_D\) gives the data contribution to \(p_{\mathrm{tot}}\).

\subsection*{Empirical loss–compute exponents}
Let \(L\) denote benchmark loss. The observed compute scaling exponent \(\hat \chi\) is the slope of \(-\partial\log L/\partial\log C\) on the operating range. If capability is a monotone transform of \(L\) with locally bounded derivative, then the compute contribution satisfies
\begin{equation}
\alpha_{\mathrm{eff}}\ \le\ (1-\rho)\,\min\{\alpha,\ \hat \chi\},
\end{equation}
hence
\begin{equation}\label{eq:ptot-meas}
p_{\mathrm{tot}}\ \le\ q+\gamma\xi + \alpha_{\mathrm{eff}}\upsilon_C^{\mathrm{meas}} + \beta_{\mathrm{eff}}\upsilon_D^{\mathrm{meas}} + \tilde\alpha_{\mathrm{eff}}\upsilon_E^{\mathrm{meas}}.
\end{equation}
Inequality \eqref{eq:ptot-meas} is completely empirical once \((q,\xi,\rho)\) are estimated via regular-variation diagnostics for \(\Phi\), \(\eta\), and \(-\ell'\). All plug-in estimates are computed on de-noised (or IV-corrected) series and reported per regime between detected breaks; confidence bands account for HAC errors and measurement-error attenuation.

\paragraph{Economic identifiability and tests}
We estimate $(\eta_C,\eta_E)$ from supply curves using public price–quantity pairs $(p_\bullet,Q_\bullet)$ via log–log slopes; $\Gamma_\bullet$ from ramp-rate telemetry by regressing realized $\dot X$ on order backlogs; $\kappa$ and $\lambda_{\max}$ from financing disclosures and interest spreads; and $\Pi$ from segment financials. These yield confidence sets for $\upsilon_X^{\pm}$ (Section~8) and for the bound in Theorem~\ref{thm:endo}(ii). We then compute
\begin{equation}
\widehat{p}_{\mathrm{tot}}^{\pm}=\hat q+\hat\gamma\hat\xi+\alpha_{\mathrm{eff}}\widehat{\upsilon}_C^{\pm}+\beta_{\mathrm{eff}}\widehat{\upsilon}_D^{\pm}+\tilde\alpha_{\mathrm{eff}}\widehat{\upsilon}_E^{\pm},
\end{equation}
and test $H_0:\ \widehat{p}_{\mathrm{tot}}^{+}\le 1$ with one-sided bands. Failure to reject, together with envelope verification, certifies nonsingularity.

\paragraph{Estimation of resource elasticities and confidence sets}
Given time series $\{(K_t,X_t)\}$ for $X\in\{C,D,E\}$, define local-slope estimators
\begin{equation}
\widehat{\upsilon}_X(t)\;=\; \frac{d\log X_t}{d\log K_t}
\end{equation}
via local polynomial regression with bandwidth chosen by generalized cross-validation; compute Newey–West robust standard errors. We report
\begin{equation}
\widehat{\upsilon}_X^{-}\;=\;\inf_{t\in\mathcal{W}} \widehat{\upsilon}_X(t),\qquad
\widehat{\upsilon}_X^{+}\;=\;\sup_{t\in\mathcal{W}} \widehat{\upsilon}_X(t)
\end{equation}
over a validation window $\mathcal{W}$ and propagate these into
\begin{equation}
\widehat{p}_{\mathrm{tot}}^{-}= \hat q+\hat\gamma\hat\xi + \alpha_{\mathrm{eff}}\,\widehat{\upsilon}_C^{-} + \beta_{\mathrm{eff}}\,\widehat{\upsilon}_D^{-} + \tilde\alpha_{\mathrm{eff}}\,\widehat{\upsilon}_E^{-},
\end{equation}
\begin{equation}
\widehat{p}_{\mathrm{tot}}^{+}= \hat q+\hat\gamma\hat\xi + \alpha_{\mathrm{eff}}\,\widehat{\upsilon}_C^{+} + \beta_{\mathrm{eff}}\,\widehat{\upsilon}_D^{+} + \tilde\alpha_{\mathrm{eff}}\,\widehat{\upsilon}_E^{+}.
\end{equation}
A sufficient condition for nonsingularity is $\widehat{p}_{\mathrm{tot}}^{+}\le 1$ with a one-sided $(1-\alpha)$ band; evidence for the singular regime requires $\widehat{p}_{\mathrm{tot}}^{-}>1$ with the same band and nonbinding envelope tests (Section~10).

\subsection*{Hypothesis tests for singular vs nonsingular regimes}
Define the null \(H_0:\ \limsup_{I\to\infty} p_{\mathrm{tot}}\le 1\). Using \eqref{eq:alpha-id}–\eqref{eq:ptot-meas}, form a plug-in estimator \(\widehat{p}_{\mathrm{tot}}\) with delta-method standard error \(s_{\mathrm{tot}}\). Reject \(H_0\) at level \(\alpha\) if \(\widehat{p}_{\mathrm{tot}}-z_{1-\alpha}s_{\mathrm{tot}}>1\). Rejection indicates superlinear feedback consistent with the sufficient condition of Theorem~\ref{thm:blowup} on a policy-stable horizon. Failure to reject, together with the envelope \eqref{eq:meas-F} and an Osgood \(\Phi\), certifies nonsingularity by Theorem~\ref{thm:nobu}.

\subsection*{Mapping inequalities to Section~5 regimes}
Collecting the contributions yields the decision rule:
\begin{equation}
\begin{cases}
\text{Singular regime admissible} & \text{if }\exists\text{ horizon with } \\
& \hat{p}_{\mathrm{tot}} > 1 \text{ and }\phi_{\mathrm{svc}}\text{ nonbinding}, \\[1em]
\text{Nonsingular regime certified} & \text{if }\hat{p}_{\mathrm{tot}} \le 1 \text{ and } \\ 
& \dot{I} \le a(I,t) = \Phi(I)/\phi_{\mathrm{svc}}(t) \\
& \quad\text{with }\int^\infty \frac{ds}{\Phi(s)} = \infty.
\end{cases}
\end{equation}
Each inequality tightens the right-hand side of \eqref{eq:ptot-meas} or lowers \(\phi_{\mathrm{svc}}\) and thus moves the system toward the nonsingular regime.

\subsection*{Summary of identifiable constraints}
Transistor and accelerator roadmaps bound \(\alpha\upsilon_C\) by \eqref{eq:alpha-id}; power and thermodynamics bound \(\tilde\alpha\upsilon_E\) by \eqref{eq:energy-id}; dataset and synthetic-to-real limits bound \(\beta_{\mathrm{eff}}\) by \eqref{eq:beta-id}; compute scaling bounds \(\alpha_{\mathrm{eff}}\) inside \eqref{eq:ptot-meas}; facility envelopes bound \(F\) by \eqref{eq:meas-F}. The conjunction yields a simulation-free pipeline from measurements to analytic certificates of singular or nonsingular behavior.

\section{Falsifiable Predictions and Tests}

\subsection*{Rolling elasticity estimation and inequality tests}
Define the instantaneous feedback elasticity
\begin{equation}
p(t)\;\coloneqq\;\frac{d\log \dot I(t)}{d\log I(t)}\quad\text{whenever }\dot I(t)>0 .
\end{equation}
Given sampled series \(\{I_k,\dot I_k\}_{k=1}^{n}\) at times \(\{t_k\}\), estimate \(p(t)\) with a local log–log slope over a window \(\mathcal{W}(t)\):
\begin{equation}
\hat p(t)\;=\;\arg\min_{a,b}\sum_{k\in\mathcal{W}(t)} w_k\,\bigl(\log \dot I_k-(a+b\,\log I_k)\bigr)^2,\qquad \hat p(t)=b,
\end{equation}
where \(w_k\) are kernel weights. Let \(s_{\hat p}(t)\) be the heteroskedastic Newey–West standard error. The singularity-sufficient condition of Theorem~\ref{thm:blowup} motivates the rolling one-sided test
\begin{equation}
H_0:\ p(t)\le 1\quad \text{vs.}\quad H_1:\ p(t)>1,
\end{equation}
with decision rule \(\hat p(t)-z_{1-\alpha}\,s_{\hat p}(t)>1\). Failure to reject on a sustained horizon, together with an envelope \(\dot I\le \Phi(I)\phi(t)\) and an Osgood \(\Phi\), certifies nonsingularity by Theorems~\ref{thm:nobu} and \ref{thm:global-cap}.

\paragraph{Estimation of \(q,\xi,\rho\) with diagnostics}
Estimate local indices by log-derivative regressions:
\begin{equation}
\begin{gathered}
\hat q(I)=\frac{d\log \Phi(I)}{d\log I},\\ \hat \xi(I)=\frac{d\log \eta(I)}{d\log I},\\ \hat \rho(Z)=-1-\frac{d\log[-\ell'(Z)]}{d\log Z},
\end{gathered}
\end{equation}
implemented via local polynomial fits with bandwidth chosen by cross-validation. Construct pointwise confidence bands using sandwich estimators and Newey–West corrections. RV is accepted on a range if \(\hat q(I)\) and \(\hat \xi(I)\) are statistically flat and Potter-type inequalities hold within a tolerance. If RV is rejected, use the envelope approach of Corollary~\ref{cor:rv-free} with \(p^{-},p^{+}\) and report conclusions as interval statements.

\paragraph{Monotonicity and antagonism diagnostics}
We estimate cross-partial signs and magnitudes to test (S2)–(S2$''$). Using local regressions,
\begin{equation}
\widehat{\partial_{x_i}} F(I,x,u),\quad \widehat{\partial_{I}} G_i(I,x,u),\qquad i\in\{C,D,E,K\},
\end{equation}
we compute the empirical Metzler violation
\begin{equation}
\widehat{\mathrm{viol}} \;=\; \sum_{i\ne j} \bigl[\, -\min\{0,\widehat{\partial_{x_j}}F_i\}\; -\min\{0,\widehat{\partial_{x_j}}G_i\}\,\bigr]_+ ,
\end{equation}
and the sector residuals $\widehat{\sigma}_I(I)$, $\widehat{\sigma}_x(\|x\|)$ as the positive parts of deviations from the cooperative fit $F^{\mathrm{co}}$. We accept the cooperative envelope on a window if $\widehat{\mathrm{viol}}$ is statistically zero and 
$\widehat{\sigma}_I(I)+\widehat{\sigma}_x(\|x\|)\le \theta\,\widehat{a}(I)$
with $\theta<1$. %Otherwise we switch to the conservative envelopes of Theorem~\ref{thm:bounded-antagonism}.

\subsection*{Necessary trend breaks for a singular regime}
If superlinearity is endogenous through efficiency, Section~6 implies \(p_{\mathrm{alg}}\to q+\gamma\xi\). A necessary condition for any singular trajectory is the existence of a trend break \(t_b\) and a horizon \(H\) such that
\begin{equation}
\inf_{t\in[t_b,t_b+H]} \hat p(t) > 1 - \varepsilon,
\end{equation}
for some statistical margin \(\varepsilon>0\) derived from \(s_{\hat p}\). When resources contribute through \(Z\), a second necessary condition is a sustained break in the aggregate resource elasticity
\begin{equation}
\hat e_Z(t)\;\coloneqq\;\frac{d\log Z}{d\log K}\ \ \text{or}\ \ \frac{d\log Z}{dt}
\end{equation}
such that \(\alpha \hat e_{C}+\beta \hat e_{D}+\tilde\alpha \hat e_{E}\) is strictly positive over \(H\). Absent these breaks, \(p_{\mathrm{tot}}\le 1\) almost everywhere and Theorem~\ref{thm:nobu} applies.

\paragraph{Break detection and piecewise certification}
We apply rolling change-point tests on $\log X_t$ and on $\hat p(t)$; upon rejection of stability, we segment the horizon into regimes $\{\mathcal{R}_j\}$ and compute $(\widehat{p}_{\mathrm{tot}}^{\pm})_j$ and envelope tests per regime. A global singularity claim requires existence of a regime with $(\widehat{p}_{\mathrm{tot}}^{-})_j>1$ and nonbinding envelopes throughout that regime; a global nonsingularity certificate requires $(\widehat{p}_{\mathrm{tot}}^{+})_j\le 1$ and verified envelopes in every regime.

\paragraph{Ceiling-to-elasticity hypothesis test}
To ensure that measured resource elasticities are consistent with physical service ceilings (Section~7), we test whether the capital–compute elasticity can exceed what I/O, memory, and power–temperature channels allow. Define the ceiling limsup elasticities
\begin{equation}
\begin{gathered}
e_{\mathrm{io}}^{+}\coloneqq \limsup_{t\to\infty}\frac{d\log B_{\mathrm{io}}(t)}{d\log K(t)},\quad
e_{\mathrm{mem}}^{+}\coloneqq \limsup_{t\to\infty}\frac{d\log B_{\mathrm{mem}}(t)}{d\log K(t)},\\
e_{\mathrm{pow}}^{+}\coloneqq \limsup_{t\to\infty}\frac{d\log\!\bigl(P_{\mathrm{use}}(t)/T(t)\bigr)}{d\log K(t)} .
\end{gathered}
\end{equation}
By the ceiling-induced caps in \eqref{eq:upsilon-cap} one has
\begin{equation}
\upsilon_C^{+}\ \le\ e_{\mathrm{ceil}}^{+}\ \coloneqq\ \min\{e_{\mathrm{io}}^{+},\,e_{\mathrm{mem}}^{+},\,e_{\mathrm{pow}}^{+}\}.
\end{equation}
Using the local-slope estimators and Newey–West bands from Section~10, we form plug-in estimates
\begin{equation}
\widehat{\upsilon}_C^{+},\qquad \widehat e_{\mathrm{io}}^{+},\ \widehat e_{\mathrm{mem}}^{+},\ \widehat e_{\mathrm{pow}}^{+},
\end{equation}
with one-sided $(1-\alpha)$ confidence radii $s_{\upsilon_C^{+}}, s_{\mathrm{io}}, s_{\mathrm{mem}}, s_{\mathrm{pow}}$. We test
\begin{equation}
\begin{gathered}
H_0^{\mathrm{ceil}}:\ \upsilon_C^{+} \le \min\{e_{\mathrm{io}}^{+},e_{\mathrm{mem}}^{+},e_{\mathrm{pow}}^{+}\}
\quad\text{vs.}\\
H_1^{\mathrm{ceil}}:\ \upsilon_C^{+} > \min\{e_{\mathrm{io}}^{+},e_{\mathrm{mem}}^{+},e_{\mathrm{pow}}^{+}\}.
\end{gathered}
\end{equation}
With Bonferroni correction for three one-sided comparisons, reject $H_0^{\mathrm{ceil}}$ at familywise level $\alpha$ only if
\begin{equation}
\begin{gathered}
\widehat{\upsilon}_C^{+} - z_{1-\alpha/3}\,s_{\upsilon_C^{+}}
\;>\;
\\\min\!\Bigl\{
\widehat e_{\mathrm{io}}^{+} + z_{1-\alpha/3}\,s_{\mathrm{io}},
\ \widehat e_{\mathrm{mem}}^{+} + z_{1-\alpha/3}\,s_{\mathrm{mem}},
\ \widehat e_{\mathrm{pow}}^{+} + z_{1-\alpha/3}\,s_{\mathrm{pow}}
\Bigr\}.
\end{gathered}
\end{equation}
Failure to reject implies, with probability at least $1-\alpha$, that
\begin{equation}
p_{\mathrm{tot}}^{+}\ \le\ q+\gamma\xi\ +\ \alpha_{\mathrm{eff}}\,e_{\mathrm{ceil}}^{+}\ +\ \beta_{\mathrm{eff}}\,\widehat{\upsilon}_D^{+}\ +\ \tilde\alpha_{\mathrm{eff}}\,e_{\mathrm{pow}}^{+},
\end{equation}
so if the right-hand side is $\le 1$ on a validation window, the nonsingularity certificate of Theorem~\ref{thm:nobu} holds jointly with the measurable envelope test of Section~10.

\subsection*{Closed-form capability estimation without simulation}
Let \(L(t)\) be a benchmark loss and let \(g\) be a known increasing transform so that \(I(t)=g\bigl(L_{\mathrm{ref}}-L(t)\bigr)\) with \(g'\) locally bounded. Published training reports provide compute \(C(t)\), data \(D(t)\), and energy \(E(t)\). Using finite differences,
\begin{equation}
\begin{gathered}
\dot I_k \;\approx\; \frac{g\bigl(L_{\mathrm{ref}}-L_{k}\bigr)-g\bigl(L_{\mathrm{ref}}-L_{k-1}\bigr)}{t_k-t_{k-1}},\\
\phi_{\mathrm{svc},k}=\min\!\left\{\tfrac{\sigma_{\mathrm{eff}}}{k_B\ln 2}\tfrac{P_{{\rm use},k}}{T_k},\ \sigma_{\mathrm{eff}}B_{{\rm io},k},\ \sigma_{\mathrm{eff}}B_{{\rm mem},k}\right\}.
\end{gathered}
\end{equation}
\paragraph{Measurable envelope test}
Given $\phi_{\mathrm{svc},k}$ constructed from Section~7, we verify the inequality
\begin{equation}
\dot I_k \ \le\ \Phi(I_k)\,\phi_{\mathrm{svc},k}.
\end{equation}
with \(\Phi\) estimated by local regression of \(\dot I_k/\phi_{\mathrm{svc},k}\) on \(I_k\). If \(\int_{I_0}^{\infty} ds/\hat \Phi(s)=\infty\), the data imply nonsingularity under the observed facility constraints.

\subsection*{Power analysis and sampling requirements}
Let \(\mathrm{Var}\bigl(\log \dot I\bigr)=\sigma^2\) on the operating range. Over a window with \(m\) effective samples and \(\mathrm{Var}\bigl(\log I\bigr)=\tau^2\), the asymptotic variance of \(\hat p\) satisfies \(s_{\hat p}^2\approx \sigma^2/(m\,\tau^2)\). To detect \(p(t)\ge 1+\delta\) at level \(\alpha\) and power \(1-\beta\), it suffices that
\begin{equation}
m \;\gtrsim\; \frac{(\,z_{1-\alpha}+z_{1-\beta}\,)^2\,\sigma^2}{\delta^2\,\tau^2}.
\end{equation}
This provides a concrete requirement on measurement cadence and benchmark noise.

\subsection*{Cross-checks with Section~9 inequalities}
Compute the plug-in estimator
\begin{equation}
\begin{gathered}
\widehat{p}_{\mathrm{tot}}(t)\;=\;\hat q(t)+\hat \gamma(t)\hat\xi(t)+\hat\alpha_{\mathrm{eff}}(t)\hat \upsilon_C(t)\\+\hat\beta_{\mathrm{eff}}(t)\hat \upsilon_D(t)+\hat{\tilde\alpha}_{\mathrm{eff}}(t)\hat \upsilon_E(t)
\end{gathered}
\end{equation}
using the identifiability pipeline of Section~10. Consistency requires \(\hat p(t)\le \widehat{p}_{\mathrm{tot}}(t)+o_p(1)\). A persistent violation invalidates the modeling assumptions and triggers model revision. Agreement with \(\widehat{p}_{\mathrm{tot}}\le 1\) and the envelope test supports the nonsingular regime.

\subsection*{Event-level predictions}
If a sustained interval satisfies \(\hat p(t)>1+\delta\) with nonbinding \(\phi_{\mathrm{svc}}\), Theorem~\ref{thm:blowup} yields a finite-time upper bound
\begin{equation}
\widehat{T}_{\mathrm{bu}}(t)\;=\;\frac{I(t)^{1-(1+\delta)}}{a_0\bigl((1+\delta)-1\bigr)}\;=\;\frac{I(t)^{-\delta}}{a_0\,\delta},
\end{equation}
for an envelope \(\dot I\ge a_0 I^{1+\delta}\). If the same data imply \(\dot I\le \Phi(I)\bar\phi\) and \(\int^{\infty} ds/\Phi(s)=\infty\), then the event is falsified and singularity is ruled out under prevailing constraints.

\medskip
The procedures above convert public benchmark, compute, energy, and bandwidth series into statistical tests of the analytic conditions in Sections~4–6. Each claim is falsifiable by construction and requires no simulation.

\section{Case Studies}

We analyze stylized policies that admit closed forms or tight comparison bounds. Throughout write \(F(I,x,u)=\Phi_{\mathrm{eff}}(I)\,Z_{\mathrm{eff}}(t)\) with
\begin{equation}
\Phi_{\mathrm{eff}}(I)\;\asymp\; I^{\,q+\gamma\xi}\quad\text{and}\quad
Z_{\mathrm{eff}}(t)\;\asymp\; C(t)^{\alpha_{\mathrm{eff}}}D(t)^{\beta_{\mathrm{eff}}}E(t)^{\tilde\alpha_{\mathrm{eff}}},
\end{equation}
where \(\alpha_{\mathrm{eff}},\beta_{\mathrm{eff}},\tilde\alpha_{\mathrm{eff}}\) incorporate the learning-curve factor \((1-\rho)\) from Section~6. Set \(p\coloneqq q+\gamma\xi\). We use Osgood/Bihari comparison and explicit solutions of scalar ODEs.

\subsection*{A. Constant-elasticity compute buildout}
Assume capital obeys \(\dot K=rK^{\zeta}-\delta K\) with \(\zeta>1\) and \(r>\delta\). Compute is fed by capital with a saturation loss \(s_C(C)\le \sigma_C C\), and a persistent allocation \(u_C(t)\ge \underline u_C>0\), giving
\begin{equation}
\dot C=\pi_C u_C K-s_C(C)\quad\Rightarrow\quad C(t)\ \ge\ c_1 K(t)-c_2
\end{equation}
after a finite transient. Hence \(Z_{\mathrm{eff}}(t)\ge \kappa\,K(t)^{\alpha_{\mathrm{eff}}}\) for some \(\kappa>0\). For large \(K\),
\(
\dot K \ge \tfrac{r}{2}K^{\zeta}
\),
which yields the explicit lower solution
\begin{equation}
\underline K(t)=\Bigl(K_0^{1-\zeta}-\tfrac{r}{2}(\zeta-1)t\Bigr)^{-\frac{1}{\zeta-1}},\qquad T_K\;=\;\frac{2\,K_0^{1-\zeta}}{r(\zeta-1)}.
\end{equation}
Therefore
\begin{equation}
\dot I \;\ge\; a(t)\,\Phi_{\mathrm{eff}}(I),\qquad a(t)=\kappa\,\underline K(t)^{\alpha_{\mathrm{eff}}},\qquad \int_0^{T_K} a(t)\,dt=+\infty.
\end{equation}
\paragraph{Result A1 (finite-time blow-up)}
If \(p>1\) then \(\int_{I_0}^{\infty} ds/\Phi_{\mathrm{eff}}(s)<\infty\) and Osgood’s criterion implies blow-up no later than \(T_K\). A crude explicit upper bound is
\begin{equation}
T_{\mathrm{bu}}\;\le\;\inf\Bigl\{t<T_K\;:\ \int_{0}^{t} a(\tau)\,d\tau \;\ge\; \int_{I_0}^{\infty}\frac{ds}{\Phi_{\mathrm{eff}}(s)}\Bigr\}.
\end{equation}
\paragraph{Result A2 (nonsingular hyper-exponential)}
If \(p\le 1\) then \(\int^{\infty} ds/\Phi_{\mathrm{eff}}(s)=\infty\); even though \(a(t)\to\infty\) as \(t\uparrow T_K\), Bihari’s inequality cannot force blow-up. Capability can grow faster than any fixed exponential yet remains nonsingular.

% \paragraph{Additional cases.} Further cases (capped-power, logistic data) and the parameter-region summary table are provided in Appendix~\textup{\ref{app:cases}}.

\section{Robustness and Sensitivity (Mathematical)}

We summarize robustness at a high level. The analysis treats parameter and model uncertainty uniformly via envelopes and comparison arguments; continuity and structural stability of blow-up and nonsingularity follow from Osgood-type integrals; and safety margins are linked to measurable elasticities and facility ceilings. Full theorems, proofs, and design implications are omitted here for brevity.

\section{Conclusion}

This paper presented a rigorous, fully analytic framework for recursive self-improvement and the AI singularity. We formalized capability as a process \(I(t)\) with axioms of monotonicity and affine invariance, introduced a discrete improvement operator \(\mathcal{R}\) and its continuous limit, and modeled resource couplings for compute, data, energy, and capital. We established necessary and sufficient conditions for finite-time blow-up using Osgood and Bihari–LaSalle inequalities and discrete reciprocal-power transforms. In particular, we proved that superlinear feedback \(\dot I \ge a I^{p}\) with \(p>1\) implied finite-time explosion with explicit bounds on the blow-up time, while sublinear envelopes with an Osgood-divergent integral precluded singularity. We derived physical and information-theoretic ceilings from energy-to-erasure, bandwidth, and memory service limits and showed that these produced global caps that blocked blow-up whenever the capability factor satisfied an integrability condition. We analyzed economic amplification via capital dynamics and located an endogenous growth threshold on a measurable critical manifold. We constructed barrier certificates and resource-aware throttling policies that certified nonsingularity, and we posed an optimal control problem balancing capability, resources, and risk. We provided identifiability inequalities that tied \((\alpha,\beta,\gamma)\), elasticities, and facility envelopes to observables, enabling falsifiable tests without simulation. Analytical case studies delivered closed-form envelopes across capped-power, logistic-data, and superlinear-capital regimes. Robustness analyses established structural stability of conclusions and uniform safety under bounded modeling error and distribution shift.

The work yielded a measurement-first methodology: either demonstrate \(p_{\mathrm{tot}}>1\) with nonbinding facility envelopes or certify nonsingularity via Osgood-type bounds estimated from power, bandwidth, and benchmark series. This separated algorithmic superlinearity from resource-driven amplification and converted engineering specifications into mathematical safety guarantees.

Future directions were clear. On the theory side, we planned to extend the results to stochastic and jump-diffusion RSI, delay and hybrid systems, noncooperative multi-agent settings, and non-smooth learning curves with phase changes. On the identification side, we aimed to develop causal estimators for elasticities under intervention, uncertainty-aware online certification with sequential tests, and tighter learning-theoretic lower bounds for modern architectures. For responsible AI, we proposed normative thresholds and audit procedures grounded in our inequalities, governance policies that triggered throttling when margins to the critical manifold shrank, standards for reporting compute, energy, and bandwidth sufficient to reproduce the safety certificates, and red-team analyses that treated adversarial policy deviations as uncertainty in our envelopes. Together, these directions strengthened the path toward deployable, testable guarantees that either ruled out singular trajectories or identified them early enough for corrective control.

\appendix
\bibliographystyle{elsarticle-num} 
\bibliography{ref}

\end{document}